\documentclass[titlepage,oneside,12pt,letter,fleqn]{article}

\usepackage[tiny,rm]{titlesec}
\newpagestyle{trbstyle}{
	\sethead{}{}{\thepage}
}
\titleformat{\section}{\bfseries}{}{0pt}{\uppercase}
\titlespacing*{\section}{0pt}{12pt}{*0}
\titleformat{\subsection}{\bfseries}{}{0pt}{}
\titlespacing*{\subsection}{0pt}{12pt}{*0}
\titleformat{\subsubsection}{\itshape}{}{0pt}{}
\titlespacing*{\subsubsection}{0pt}{12pt}{*0}

\usepackage{enumitem}
\setlist[1]{labelindent=0.5in,leftmargin=*}
\setlist[2]{labelindent=0in,leftmargin=*}

\usepackage{ccaption}
\usepackage{amsmath}
\usepackage{amsfonts}
\usepackage{booktabs}
\usepackage{float}
\usepackage{placeins}
\makeatletter
\renewcommand{\fnum@figure}{\textbf{FIGURE~\thefigure} }
\renewcommand{\fnum@table}{\textbf{TABLE~\thetable} }
\makeatother
\captiontitlefont{\bfseries \boldmath}
\captiondelim{\;}

\usepackage[sort,numbers]{natbib}

\setcitestyle{round}

\usepackage{lineno}

\usepackage{totcount}
	\regtotcounter{table} 	
	\regtotcounter{figure} 	

\newcommand\wordcount{
    \immediate\write18{texcount -sum -1 \jobname.tex > 'count.txt'} \input{count.txt} }

\usepackage{mathptmx}
\usepackage{times}
\usepackage{cases}
\usepackage{textcomp}
\usepackage{graphicx}
\usepackage{subfig}
\usepackage{url}
\usepackage{epstopdf}
\usepackage{color}
\usepackage{mathrsfs,MnSymbol}
\usepackage{booktabs}
\usepackage{multirow}
\usepackage{soul,color}
\usepackage{enumitem}

\usepackage{cleveref}

\usepackage{caption}

\begin{document}

\thispagestyle{empty}

\begin{titlepage}
\begin{flushleft}

{\uppercase{ \textbf{Fairness-enhancing deep learning for ride-hailing demand prediction}}
}\\[1cm]

\textbf{Yunhan Zheng}\\
Department of Civil and Environmental Engineering\\
Massachusetts Institute of Technology\\
77 Massachusetts Ave, Cambridge, MA 02139\\
Email: yunhan@mit.edu\\[0.5cm]

\textbf{Qingyi Wang}\\
Department of Civil and Environmental Engineering\\
Massachusetts Institute of Technology\\
77 Massachusetts Ave, Cambridge, MA 02139\\
Email: qingyiw@mit.edu\\[0.5cm]

\textbf{Dingyi Zhuang}\\
Department of Civil and Environmental Engineering\\
Massachusetts Institute of Technology\\
77 Massachusetts Ave, Cambridge, MA 02139\\
Email: dingyi@mit.edu\\[0.5cm]

\textbf{Shenhao Wang (Corresponding Author)}\\
Department of Urban Studies and Planning\\
Massachusetts Institute of Technology\\
77 Massachusetts Ave, Cambridge, MA 02139\\
Email: shenhao@mit.edu\\[0.5cm]

\textbf{Jinhua Zhao}\\
Department of Urban Studies and Planning\\
Massachusetts Institute of Technology\\
77 Massachusetts Ave, Cambridge, MA 02139\\
Email: jinhua@mit.edu\\[0.5cm]

\end{flushleft}
\end{titlepage}

\newpage

\thispagestyle{empty}
\begin{center}
\large{\textbf{Fairness-enhancing deep learning for ride-hailing demand prediction}}\\[1cm]
\end{center}

\section*{Abstract}
Short-term demand forecasting for on-demand ride-hailing services is one of the fundamental issues in intelligent transportation systems. However, previous travel demand forecasting research predominantly focused on improving prediction accuracy, ignoring fairness issues such as systematic underestimations of travel demand in disadvantaged neighborhoods. This study investigates how to measure, evaluate, and enhance prediction fairness between disadvantaged and privileged communities in spatial-temporal demand forecasting of ride-hailing services. A two-pronged approach is taken to reduce the demand prediction bias. First, we develop a novel deep learning model architecture, named socially aware neural network (SA-Net), to integrate the socio-demographics and ridership information for fair demand prediction through an innovative socially-aware convolution operation. Second, we propose a bias-mitigation regularization method to mitigate the mean percentage prediction error gap between different groups. The experimental results, validated on the real-world Chicago Transportation Network Company (TNC) data, show that the de-biasing SA-Net can achieve better predictive performance in both prediction accuracy and fairness. Specifically, the SA-Net improves prediction accuracy for both the disadvantaged and privileged groups compared with the state-of-the-art models. When coupled with the bias mitigation regularization method, the de-biasing SA-Net effectively bridges the mean percentage prediction error gap between the disadvantaged and privileged groups, and also protects the disadvantaged regions against systematic underestimation of TNC demand. Our proposed de-biasing method can be adopted in many existing short-term travel demand estimation models, and can be utilized for various other spatial-temporal prediction tasks such as crime incidents predictions. This is one of the first studies to consider prediction fairness in short-term travel demand forecasting.

\bigskip
\noindent
\emph{Keywords}: Spatial-temporal travel demand prediction, algorithmic fairness, demand forecasting, ride-hailing service
\newpage

\section{1. Introduction}
In recent years, on-demand ride-hailing services have grown rapidly. Transportation network companies (TNCs) such as Uber and Lyft provide the ride-hailing services by connecting passengers with drivers based on real-time information \cite{diao2021impacts,ZHENG2023103639}. Reliable and accurate short-term travel demand forecasting is a promising tool to balance vehicle supply and demand with low cost and high quality of service \cite{ke2017short,guo2022data,guo2021robust}. Researchers have developed a series of data-driven approaches to predict travel demand in real-time, including time series analysis methods \cite{zhang2011seasonal,li2012prediction}, machine learning methods \cite{xu2013public,li2015traffic} and deep learning models \cite{ke2017short,guo2020residual,li2021multi}. These approaches typically divide the study region into small areas, use the past travel requests in a time interval as the historical demand, and then seek to enhance the prediction accuracy of the future travel demand as a function of the historical demand (assuming certain spatial and temporal correlations among them) and exogenous features such as the weather and holiday.\\

\noindent  However, a narrow focus on prediction accuracy ignores the crucial social consequences underlying the prediction tasks, such as unfairness in travel demand forecasting. For instance, since the transport operators depend on the predicted passenger demand to dispatch vehicles, systematically under-predicted travel demand in disadvantaged neighborhoods may lead to inadequate service provision for certain groups. The existing literature has the following two limitations: first, most previous studies evaluated the performance of the demand predictions by the average prediction accuracy across the whole study region, while research into the disparity of predictive performance between the disadvantaged and privileged areas is very scarce. This raises an equity concern because if the ride-hailing demand for the disadvantaged neighborhoods is systematically underestimated, the vehicles allocated to these neighborhoods may not be enough to serve the actual demand. Second, most previous models did not consider the socioeconomic and demographic information of the areas when making travel demand predictions. Areas with different socioeconomic and demographic makeup could have very different spatial-temporal dependencies. Failure to account for the heterogeneity of these spatial-temporal dependencies can lead to biased model results. \\

\noindent To overcome these limitations, this paper proposes a novel strategy to improve prediction fairness while retaining high prediction accuracy. This strategy is comprised of a new deep learning architecture, named the \textit{socially aware network} (SA-Net), and a bias-mitigation regularization method, to achieve fairness-aware travel demand predictions. While previous research typically adopted spatially-invariant convolutional kernels to capture spatial dependencies, this new network incorporates a novel Socially-Aware Convolution (SAC) module that adapts the standard invariant kernel at each area of the study region based on the socio-demographic makeup of that area, which is highly flexible and thus can better capture the spatial-temporal dependencies across different locations. The bias-mitigation regularization method modifies the traditional objective functions in deep learning travel demand predictions by adding a fairness regularization term, thus facilitating fair travel demand predictions.\\

\noindent To the best of our knowledge, this paper is one of the first attempts to improve prediction fairness in spatial-temporal travel demand forecasting of on-demand ride-hailing service. The main contributions of this paper can be summarised as follows:
\begin{itemize}
\item We propose a new model (SA-Net) that adopts location-specific modification to the standard spatially-invariant convolutional filters. The proposed network can flexibly capture the spatial heterogeneity by incorporating the local socioeconomic and demographic information into the prediction process.
\item We propose a fairness metric, the \textit{mean percentage error gap} (MPE Gap), which measures the gap of mean percentage prediction error between the disadvantaged and privileged groups. A positive MPE indicates that the model has underestimated the demand, whereas a negative MPE indicates an overestimation of the demand. 
\item We develop a bias-mitigation regularization method that allows the network to learn fair predictions by bringing down the MPE Gap between the disadvantaged and privileged groups.
\item Experiments on Chicago TNC data reveal the risk of generating spatially unfair demand prediction with the state-of-the-art spatial-temporal deep learning predictions, and show that our proposed new method can not only reduce the fairness gap between the disadvantaged and privileged groups, but also increase the overall prediction accuracy.
\end{itemize}

\noindent The rest of the paper is organized as follows. Section \ref{sec:literature} reviews the existing literature on ride-hailing demand prediction and fairness in machine learning. Section \ref{sec:problem} defines the research problem. Section \ref{sec:method} describes the model architecture of the proposed SA-Net, the fairness evaluation metrics, as well as the bias mitigation regularization method. Section \ref{sec:experiment} shows the experiment results, which compare the prediction accuracy and fairness between the proposed de-biasing SA-Net and the benchmark models on the Chicago TNC dataset. Section \ref{sec:conclusion} concludes the paper.


\section{2. Literature Review}
\label{sec:literature}
\subsection{2.1 Spatial-temporal travel demand forecasting}
Short-term travel demand forecasting has been a fundamental issue in intelligent transportation systems \cite{yang2020using,ke2017short,guo2020residual,wu2020inductive,liu2022universal}. The earliest research on travel demand forecasting was based on traditional time-series regression models such as Autoregressive Integrated Moving Average model (ARIMA) \cite{zhang2011seasonal}, Kalman Filter and their variants. For example, \citet{li2012prediction} proposed an improved ARIMA based prediction method to forecast the spatial-temporal variation of passengers in urban hotspots. \citet{lippi2013short} coupled the Seasonal ARIMA (SARIMA) model with Kalman filter to capture the seasonal patterns of temporal information for freeway traffic flow prediction.\\

\noindent In recent years, researchers are moving from classical statistical models to machine learning based approaches because of the explosive growth of data accessibility and computing power  \cite{vlahogianni2014short}. Early machine learning models include support vector regression \cite{xu2013public} and regression trees \cite{li2015traffic}. More recently, deep learning methods have become increasingly popular due to its capability of approximating decision functions and its interpretability through economics theory \cite{wang2020econ_info,wang2020architecture}. Studies also investigated how to integrate deep learning and classical statistics in travel demand prediction using statistical learning theory \cite{wang2021stat_learning,wang2021residual}. The network structures in deep learning can successfully capture the complicated spatial-temporal correlations in the travel demand data for prediction. The major components are summarised below. \\

\noindent Recurrent Neural Network (RNN) is one of the most frequently used deep learning architecture to capture the sequential dependencies \cite{li2020origin}. However, RNNs suffer from the ``vanishing gradient" problem which makes them impossible to learn long data sequences \cite{yu2017deep}. As such, Long Short-Term Memory (LSTM) was developed as an enhanced form of RNNs and was widely adopted to explore long-range temporal dependencies in the data \cite{fu2016using,xu2018station}.\\

\noindent Previous studies have found that vehicle accumulation and dissipation can influence the traffic volume of nearby locations, therefore spatial correlations should be considered in demand forecasting \cite{zhu2014traffic}. Convolutional Neural Networks (CNNs) have been widely adopted to capture the spatial correlations in grid-based travel demand predictions. CNNs capture the spatial dependencies using localized kernels. They were initially designed for the analysis of visual imagery, and were later applied to learn the local and global spatial correlations in travel demand forecasting \cite{yu2017deep}. Graph Neural Networks (GCNs) are capable of capturing non-Euclidean spatial correlations in network-structured data \cite{lin2018predicting,cui2019traffic,zhang2019multistep}. GCNs were mostly applied in the station-based or traffic
network accessible scenarios, and have only recently been applied to the region-based scenarios \cite{jin2020urban}.\\

\noindent Some recent studies have integrated different deep learning neural networks to account for the complex spatial-temporal relationships in travel demand. For instance, \citet{ke2017short} developed a Conv-LSTM network for short-term passenger demand forecasting which combines CNN and LSTM to capture spatial, temporal, and exogenous dependencies simultaneously. \citet{guo2020residual} combined an extended Conv-LSTM and fully convolutional neural networks via residual connections. \citet{ke2021predicting} combined a residual multi-graph convolutional (RMGC) network and a LSTM network to improve the travel demand prediction performance.\\

\noindent Although the abovementioned methods have made remarkable progress on improving the prediction accuracy, most of them do not consider fairness when making predictions. Fairness essentially involves the evaluation of a predictive model regarding its social consequences, so without incorporating any socio-demographic information, the predictive models are hardly aware of the social consequences. Motivated by this research gap, this paper aims to evaluate and improve fairness in TNC travel demand forecasting by incorporating socio-demographics, proposing fairness metrics, and developing an fairness-enhancing prediction method. 

\subsection{2.2 Fairness in machine learning}
There exists extensive machine learning literature showing that a model can act discriminatorily on one population in a variety of settings such as criminal risk assessment \cite{Angwin16,chouldechova2017fair}, clinical care \cite{Rajkomar18,goodman2018machine} and credit risk evaluation \cite{deku2016access,lee2019context}. These studies made significant contributions in terms of formalizing fairness in machine learning \cite{DBLP:journals/corr/abs-1808-00023,DBLP:journals/corr/abs-1710-03184}, designing fairness-enhancing algorithms \cite{friedler_comparative_2019,6137441,zhang_mitigating_2018} and solving fairness concerns in real-world industries \cite{kenneth_holstein_improving_2019,gunduz_machine_2019}. \\

\noindent However, literature that investigated the algorithmic fairness issue in transportation research was very scarce. In the domain of travel behavior modeling, \citet{zheng2021equality} demonstrated prediction disparities regarding race, income, medical condition and region in travel behavior modeling using the 2017 National Household Travel Survey (NHTS) and the 2018-2019 My Daily Travel Survey in Chicago. The authors adopted an absolute correlation regularization method to mitigate the prediction biases. In the spatial-temporal travel demand modeling setting, to the best of our knowledge, very few studies tried to tackle the fairness issue. \citet{yan2020fairness} modified the loss function in deep learning to reduce the gap of per capita predicted bikeshare demand between the disadvantaged and advantaged regions. The modification is based on the fairness assumption that the per capita predicted demand should be the same across regions.  \citet{yan2021equitensors} leveraged adversarial learning to mitigate the gap in prediction errors of bikeshare demand between the advantaged and disadvantaged groups.\\

\noindent Although much progress has been made in addressing algorithmic bias, there are still several research limitations that need to be addressed. First, one critical source of bias is feature selection, where selected variables fail to capture sufficient details that affect different outcomes \cite{barocas2016big,mehrabi2021survey}. To combat this, it is crucial to develop strategies to integrate sociodemographic information into the modeling process. Another limitation is that previous research on fairness has measured it based on the absolute value of demand, which can lead to errors in disadvantaged groups being considered insignificant. To address this, we propose a new measure of fairness based on the relative value of demand. This measure compares errors with the typical demand of the region, which is based on the concept of algorithmic fairness known as ``equality of odds''\cite{hardt2016equality}. This principle requires that all individuals who have a TNC demand should have an equal chance of having it reflected in the prediction, regardless of their social and demographic characteristics. By using this new measurement of fairness, we can better understand and mitigate algorithmic bias in ride-sharing platforms. To address these limitations, we build upon the state-of-the-art spatial-temporal travel demand models, and propose a novel method for fair predictions of TNC travel demand.

\section{3. Problem Description and Preliminaries}
\label{sec:problem}
The goal of this study is to predict the short-term TNC demand in the study area. Based on the method proposed by \citet{ke2017short} and \citet{guo2020residual}, the study area is partitioned into $I \times J$ grids with each grid referring to a zone. The temporal dimension considered is 1 hour. It is assumed that future TNC demand is correlated with the TNC demand in the past. It is also influenced by seasonality (time-of-day, day-of-week, etc.), and exogenous variables such as the weather. The variables examined in this study are defined as follows:

\begin{enumerate}[leftmargin=*]
    \item TNC demand \\
    The TNC demand at the $t$th time slot across the whole region is denoted as $\mathcal{D}_{t} \in R^{I*J}$, which is defined as the number of TNC orders happened during that time interval. The TNC demand in grid $(i,j)$ is then denoted as $(\mathcal{D}_{t})_{i,j}$
    \item Time-of-day, day-of-week, holiday\\
    By examining the Chicago traffic index data\footnote{\url{https://www.tomtom.com/en_gb/traffic-index/chicago-traffic/}}, we categorize 24h in each day into three periods: the peak hours (7am - 9am and 3pm - 7pm in workdays), the mid-peak hours (9am -3pm in workdays and 11am - 7pm in weekends), the off-peak hours (7pm - 7am in workdays and 7pm - 11am in weekends). We use $tod_{t}$ to indicate this time-of-day variable, which takes values 0, 1, 2 if $t$ belongs to the off-peak hours, the mid-peak hours and the peak hours, respectively. $dow_t$ is the day-of-week variable, which takes value 1 if $t$ is among the weekdays and 0 if $t$ is among the weekends. The dummy variable $h_t$ is used to indicate whether $t$ is in a holiday or not.
    \item Weather\\
    We consider precipitation as the weather variable, which is denoted as $p_t$. The precipitation data is obtained from the website of National Centers for Environmental Information \cite{national2021climate}.
    \item Socio-demographic data\\
    The 2019 American Community Survey (ACS) 5-year estimates data is used to produce socio-demographic data by census tract. The socio-demographic variables include total population, population per squared kilometers, percentage of African-American population, percentage of female population, percentage of spanish speakers, percentage of foreign-born population, median household income, percentage of population with 2019 household income lower than \$25,000, percentage of college graduates, percentage of population with age between 25 and 34, percentage of population with age over 65, percentage of transit commuters and percentage of population with no household vehicles. We use $Z^p$ to represent the $p^{th}$ socio-demographic variable and use $P$ to denote the total number of socio-demographic variables.
\end{enumerate}
The target of this study is to predict the TNC demand at time $t$ ($\mathcal{D}_{t}$), given the historical TNC demand, the time series features and the socio-demographic variables: \{$\mathcal{D}_{s}, p_s| s=0,...,t-1$\}, \{$tod_{s}, dow_{s}, h_s| s=0,...,t $\} and \{$Z^p| p=1,...,P $\}. This research focuses on two objectives: prediction accuracy and fairness. Prediction accuracy refers to the goal of minimizing the overall prediction errors. Prediction fairness refers to the goal of reducing the gap in mean percentage errors between the disadvantaged and privileged groups. Prediction fairness is also enhanced if the model increases the accuracy for the disadvantaged group more than the privileged group.

\section{4. Methodology}
\label{sec:method}
This research designs a novel SA-Net to predict the short-term TNC demand with enhanced fairness. We first introduce the Socially-Aware Convolution (SAC), a base module that is repeatedly used in SA-Net, and describe how SAC is adapted from the standard CNN. We will then introduce SAC-LSTM which combines SAC and LSTM. After that, we will explain the complete model architecture used in this study.
\label{sec:method}
\subsection{4.1 CNN and SAC}
\begin{figure}[!t]
    \centering
    \includegraphics[width=5.3 in]{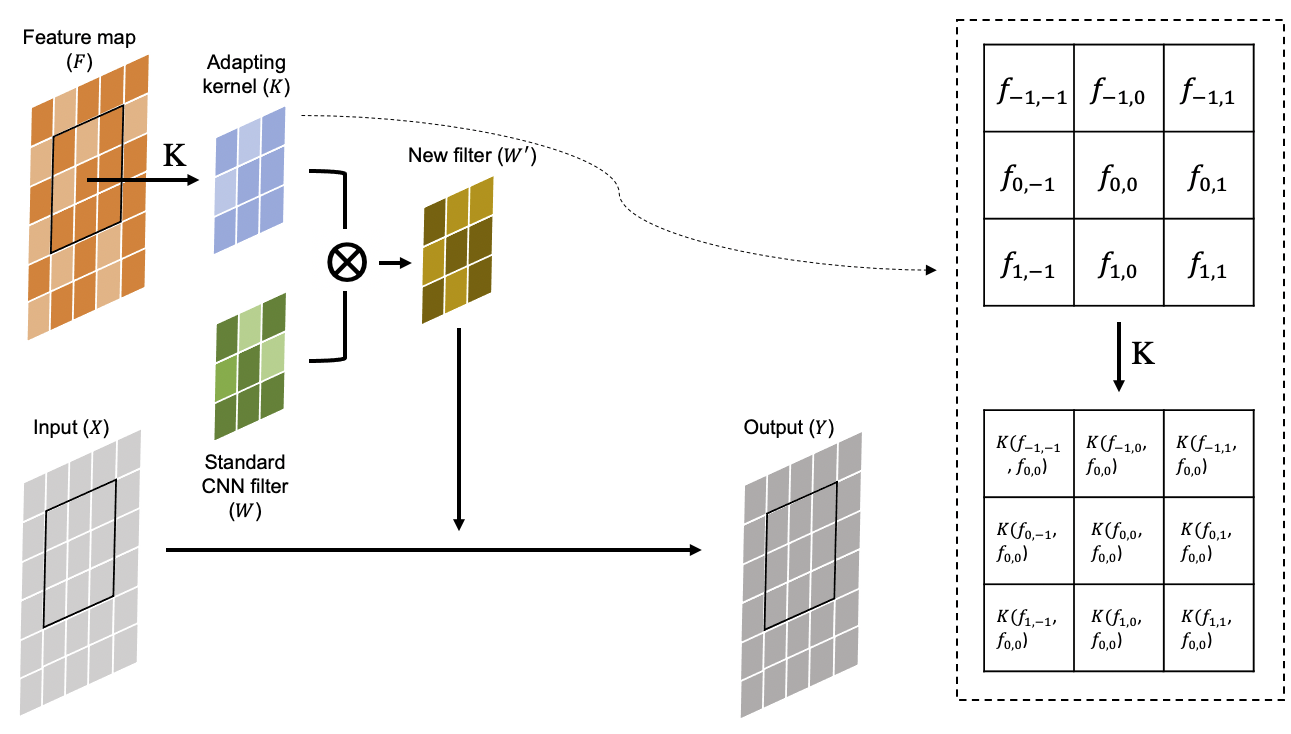}
    \caption{Socially-Aware Convolution \cite{su2019pixel}}
    \label{fig:sac}
\end{figure}

In this section, we will start with a formulation of the standard convolution neural network, and then extend it to the Socially-Aware Convolution (SAC). The concept of SAC is illustrated in Figure \ref{fig:sac}. We start from a standard convolution, which can be written as:
\begin{linenomath}
\begin{equation}
\label{eq_cnn}
    Y[m,p,q] = \sigma (\sum_{i,j,n}^{} W[m,n,i,j]*X[n,p+\hat{i},q+\hat{j}]),
\end{equation}
\end{linenomath}
where $Y \in R^{O \times S \times S}$ denotes the output tensor, $X \in R^{I \times S \times S}$ is the input tensor, $W \in R^{O \times I \times S \times S}$ denotes the filter weight. $O$, $I$, $S$ and $V$ represent the output channel size, input channel size, image size, and kernel size. $[p,q]$ denotes the pixel coordinates. $m$ and $n$ are the indices for the output and input channels. $\hat{i} = i - [V/2]$, $\hat{j} = j - [V/2]$. $\sigma$ is the activation function. From Equation \ref{eq_cnn}, we can see that the filter weight $W[m,n,i,j]$ is invariant to image locations. Therefore, the standard convolution is content-agnostic. To account for the local information, we use the Socially-Aware Convolution (SAC) which was built upon the work by \cite{su2019pixel}. A SAC modifies the spatially invariant filter $W$ with an adapting kernel $K$, which can be expressed as follows:
\begin{linenomath}
\begin{equation}
\label{eq_sac}
    Y[m,p,q] = \sigma (\sum_{i,j,n}^{} K(F[p+\hat{i},q+\hat{j}],F[p,q])*W[m,n,i,j]*X[n,p+\hat{i},q+\hat{j}]),
\end{equation}
\end{linenomath}
where $F \in R^{S \times S}$ is the feature map, which will be explained in the following subsection. $K$ represents the Gaussian kernel function: $K(f_1,f_2)=exp(-\frac{1}{2}(f_1-f_2)^T(f_1-f_2))$. The kernel values are higher for regions with similar feature values. The SAC operation represented by Equation \ref{eq_sac} adapts the standard convolution filter $W$ at each pixel by multiplying the spatially-invariant filter $W$ with a spatially-varying adapting filter $K$. The feature map $F$ picks up local features that reflect the relationships between different regions on the map. 

\subsection{4.2 Feature map construction}
\begin{figure}[!t]
    \centering
    \includegraphics[width=5.3 in]{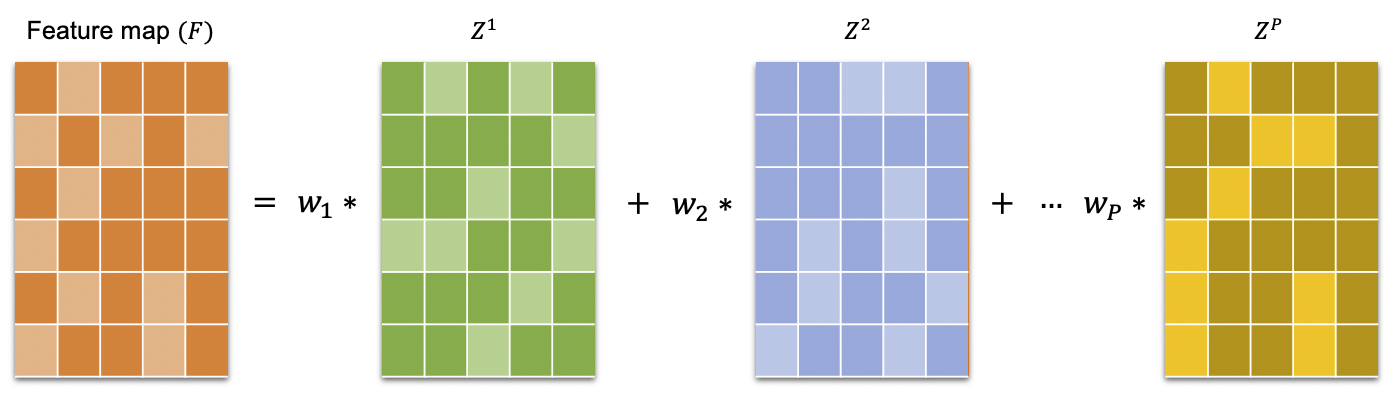}
    \caption{Feature map ($F$) construction. $Z_{ij}^p$ represents the value of the socio-demographic variable $Z^p$ for pixel [i,j]}
    \label{fig:feature_map}
\end{figure}

We construct the feature map $F$ as a linear combination of various socio-demographic variables, which is shown in Figure \ref{fig:feature_map}. The feature value $f_{ij}$ is calculated as $f_{ij} = \sum_{p}^{P}w_p*Z_{ij}^p$ where $Z_{ij}^p$ represents the value of the socio-demographic variable $p$ (e.g. population density, race, income etc.) for region $[i, j]$. By applying a Gaussian kernel function to the feature values of the center pixel and its surrounding pixels, for each pixel value prediction, we emphasize the neighboring pixels that are more similar to this specific pixel in terms of the socio-demographic features. The underlying assumption is that the regions that have similar socio-demographic characteristics with their neighborhoods should have similar level of TNC demand with their neighborhoods as well. 

\subsection{4.3 LSTM and SAC-LSTM}
We use LSTM, a special kind of Recurrent Neural Netowrk (RNN), to process the temporal information. LSTM is designed to avoid the long-term memory problem. The model first passes a sequence of input vectors to the memory cell tensors through the input gate, and then drops the redundant information through the forgot gate, and the cell state will be updated accordingly. Finally, after several iterations, the output gate will output a hidden sequence \cite{yao2018deep}. \\


\noindent When dealing with the travel demand forecasting problem with spatial-temporal data, \citet{ke2017short} proposed using the Conv-LSTM, which is a network that combines CNN and LSTM, to capture the spatial dependencies. Unlike LSTM, Conv-LSTM converts all the inputs, memory cell values, hidden states and various gates from 2D matrices to 3D tensors. Besides, Conv-LSTM replaces the Handamard product with the convolutional operator, which is used to explore spatially local correlations. However, Conv-LSTM utilizes the standard convolutional filters which are replicated across the tensors with shared weights, thus failing to account for the heterogeneity of spatial correlations. To address this drawback of standard convolutions, we modify the Conv-LSTM by replacing the standard convoluations with the SAC, and name the new network SAC-LSTM. The formulation of SAC-LSTM is as follows:
\begin{linenomath}
\begin{equation}
\label{eq_sac_lstm}
\begin{aligned}
    &\mathcal{I}_{t}=\sigma(\textit{W}_{xi}\mbox{*}\mathcal{X}_{t}+\textit{W}_{hi}\mbox{*}\mathcal{H}_{t-1}+\textit{W}_{ci}\circ \mathcal{C}_{t-1}+b_{i}) \\
    &\mathcal{F}_{t}=\sigma(\textit{W}_{xf}\mbox{*}\mathcal{X}_{t}+\textit{W}_{hf}\mbox{*}\mathcal{H}_{t-1}+\textit{W}_{cf}\circ \mathcal{C}_{t-1}+b_{f}) \\
    &\mathcal{C}_{t}=\mathcal{F}_{t}\circ \mathcal{C}_{t-1}+\mathcal{I}_{t}\circ tanh(\textit{W}_{xc}\mathcal{X}_{t}+\textit{W}_{hc}\mathcal{H}_{t-1}+b_{c})\\
    &\mathcal{O}_{t}=\sigma(\textit{W}_{xo}\mathcal{X}_{t}+\textit{W}_{ho}\mathcal{H}_{t-1}+\textit{W}_{co}\circ \mathcal{C}_{t}+b_{o})\\
    &\mathcal{H}_{t}=\mathcal{O}_{t}\circ tanh(\mathcal{C}_{t})
\end{aligned}
\end{equation}
\end{linenomath}
where the weight matrices $\textit{W}_{xf}$,$\textit{W}_{hf}$,$\textit{W}_{xc}$,$\textit{W}_{hc}$,$\textit{W}_{xo}$,$\textit{W}_{ho}$ denote the SAC weights, which are represented by $\textit{W'}$ in Figure \ref{fig:sac}. ``$\mbox{*}$'' stands for the convolutional operator. $\mathcal{I}_{t}$, $\mathcal{F}_{t}$, $\mathcal{C}_{t}$, $\mathcal{O}_{t}$, $\mathcal{H}_{t}$ are improved input gate, forgot gate, cell state, output gate and hidden state that embeds the spatial dependencies. $\circ$ denotes Hadamard product (i.e. element-wise product). $\sigma$ and $tanh$ are nonlinear activation functions:
\begin{linenomath}
\begin{equation}
\label{eq_act}
\begin{aligned}
    &\sigma(x)=\frac{1}{1+e^{-x}}; \hspace{0.2cm} tanh(x)=\frac{e^{x}-e^{-x}}{e^{x}+e^{-x}}
\end{aligned}
\end{equation}
\end{linenomath}
\subsection{4.4 Model description}
\begin{figure}[!t]
    \centering
    \includegraphics[width=4 in]{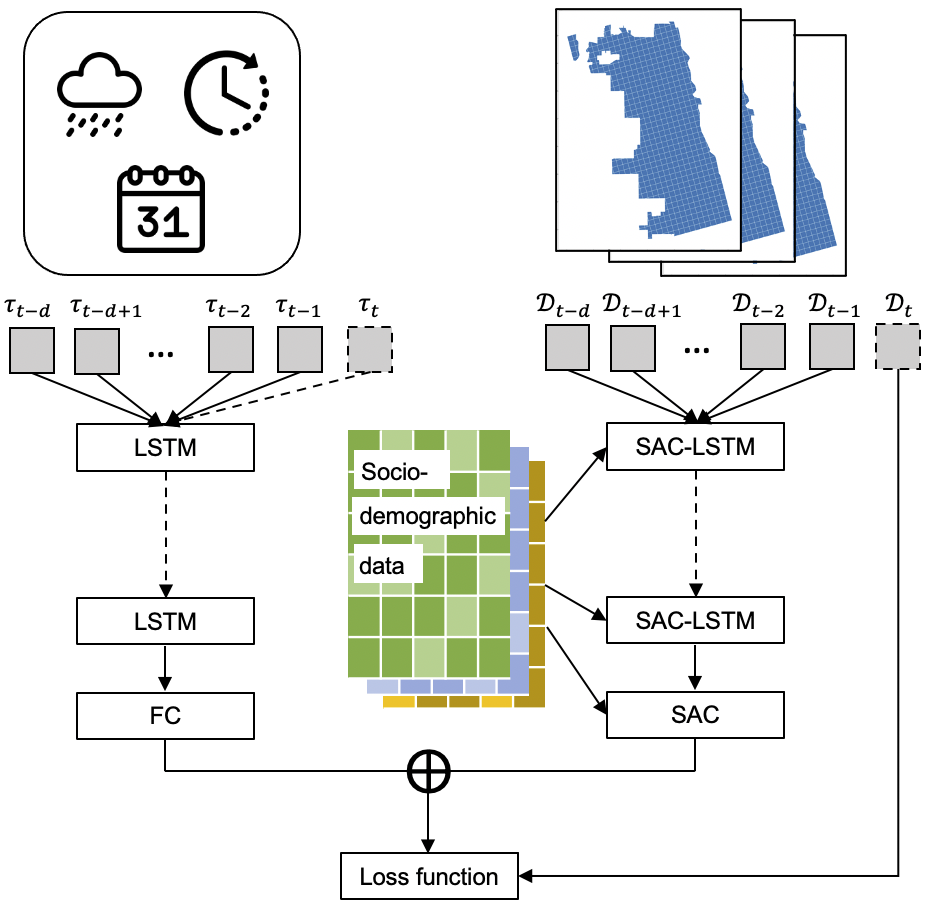}
    \caption{The structure of SA-Net}
    \label{fig:model}
\end{figure}

In this section, we propose a novel \textit{socially aware network} (SA-Net) to forecast the short-term TNC demand. The architecture of the network is illustrated in Figure \ref{fig:model}, which is comprised of two parts: the part on the right captures the spatial-temporal variables (i.e. TNC demand) using a stack of SAC-LSTM layers, and the part on the left processes the non-spatial temporal variables using a stack of LSTM layers.
\subsubsection{4.4.1 Structure for spatial-temporal variables}
We use a series of stacked SAC-LSTM layers to capture the spatial dependencies and temporal correlations for the spatial-temporal variable, which is the TNC demand data in our case. $D_t$ is used to denote the TNC demand for time slot $t$. Let $\mathcal{C}$ denote a SAC-LSTM cell: $\mathcal{C}: R^{d*M*N*I} \rightarrow R^{d*M*N*O}$, where $d$ denotes the look-back time window, which refers to the number of previous hours taken as predictors for the TNC demand in each time slot. $M$ and $N$ are the dimensions of rows and columns, $I$ and $O$ represent the number of channels for the input and output feature vectors. $L$ denotes the number of stacked SAC-LSTM layers. The formulation of the model architecture that processes the TNC demand data is written as: 
\begin{linenomath}
\begin{equation}
\label{eq_tnc}
\begin{aligned}
    &(\mathcal{U}_{t-d},\mathcal{U}_{t-d+1},...,\mathcal{U}_{t-1})=\mathcal{C}_{L}...\mathcal{C}_{1}(\mathcal{D}_{t-d},\mathcal{D}_{t-d+1},...,\mathcal{D}_{t-1})\\
    &\hat{\mathcal{X}_{t}^u}=\textit{W}_{ux}\mbox{*}\mathcal{U}_{t-1}+b_{u}
\end{aligned}
\end{equation}
\end{linenomath}
where $\mathcal{U}_{t-k}$, $k=1,2,...,d$ represent the output tensors at the last layer of the stacked SAC-LSTM layers. $W_{ux}$ represents the convolutional operation with the SAC kernel, which is applied to further capture the spatial dependency at the final layer, and also to reduce the number of output channel to 1. 

\subsubsection{4.4.2 Structure for temporal variables}
The temporal predictors used in this study include the time-related variables and the weather feature. The time-related variables include time-of-day, day-of-week and holiday indicators. The weather feature is represented by the amount of precipitation. We create a new variable $v_t=(dow_{t},tod_{t},h_{t})$ that concatenates $dow_t$, $tod_t$ and $h_t$, and use $p_t$ to represent the amount of precipitation at time $t$. These temporal features are likely to impact the TNC demand across the whole region. Then the network for the time-series variables can be written as follows:
\begin{linenomath}
\begin{equation}
\label{eq_temporal}
\begin{aligned}
    &(V_{t-d},V_{t-d+1},...,V_{t-1})=\mathcal{L}_{L}...\mathcal{L}_{1}(v_{t-d},v_{t-d+1},...,v_{t-1},v_{t})\\
    &\hat{\mathcal{X}_{t}^v}=\mathcal{F}^R(\textit{w}_{vx}V_{t-1}+b_{v})\\
        &(P_{t-d},P_{t-d+1},...,P_{t-1})=\mathcal{L}_{L}...\mathcal{L}_{1}(p_{t-d},p_{t-d+1},...,p_{t-1})\\
    &\hat{\mathcal{X}_{t}^p}=\mathcal{F}^R(\textit{w}_{px}P_{t-1}+b_{p})\\
\end{aligned}
\end{equation}
\end{linenomath}
where $V_{t-d}$ and $P_{t-d}$, $k=1,2,...,d$ are the output tensors at the last layer of the stacked LSTM layers for the time variables and the precipitation variable. $\textit{w}_{vx}$ and $\textit{w}_{px}$ denote the fully connected layers following the stacked LSTM layers, which reduce the number of output channel to 1. $\mathcal{F}^R$ denotes a reshaping function that repeat a value across the space: $\mathcal{F}^R: R \rightarrow R^{M*N*1}$, where $(\mathcal{F}^R)_{m,n,1}=x$ for any $m \in (1,2,...,M), n \in (1,2,...,N)$. $\mathcal{F}^R$ is deployed to make the dimensions of the LSTM outputs $\hat{\mathcal{X}_{v}}$ and $\hat{\mathcal{X}_{p}}$ the same with the SAC-LSTM output $\hat{\mathcal{X}_{u}}$.

\subsubsection{4.4.3 Fusion}
The final estimated TNC demand at time $t$ is a weighted combination of the estimated outputs from different parts of the network, which is given by:
\begin{linenomath}
\begin{equation}
\label{eq_fuse}
\begin{aligned}
    &\hat{\mathcal{X}_{t}}=\textit{W}_{u}\circ \hat{\mathcal{X}_{t}^u}+\textit{W}_{v}\circ \hat{\mathcal{X}_{t}^v}+\textit{W}_{p}\circ \hat{\mathcal{X}_{t}^p}
\end{aligned}
\end{equation}
\end{linenomath}
\\
\subsection{4.5 Accuracy and fairness metrics}
The performance of the various models is evaluated based on two types of metrics: the accuracy metrics and the fairness metrics. Two commonly used accuracy metrics - Mean Absolute Error (MAE) and Mean Absolute Percentage Error (MAPE) - are adopted to evaluate the prediction accuracy of the models in this work. They are defined as below:

\begin{align}
    &MAE=\frac{1}{N\times T}\sum_{t=1}^{T}\sum_{i=1}^{N} |y^i_{t}-\hat{y}^i_{t}|\\
    &MAPE=\frac{1}{T}\sum_{t=1}^{T}\frac{1}{|\mathcal{N}_t|}\sum_{i\in\mathcal{N}_t}|\frac{y^i_{t}-\hat{y}^i_{t}}{y^i_{t}}|,\; \; \mathcal{N}_t = \{i: 1 \leq i \leq N, \; y^i_{t}>0.1\}
\end{align}
where $y^i_{t}$ and $\hat{y}^i_{t}$ are the real and predicted travel demands at time interval $t$ in region $i$. $T$ represents the total number of time intervals. $N$ represents the total number of regions. $\mathcal{N}_t$ denotes the set of regions with $y^i_{t}>0.1$, which is defined to guarantee that the denominator of the absolute percentage error for the regions included is not zero. \\

\noindent While MAE and MAPE have been widely utilized to measure the accuracy of the model predictions, one limitation of these two metrics is that they do not consider the directions of the errors. Given that the underestimations and overestimations of the TNC demand predictions have very different practical implications which should not be ignored, we also examine the Mean Percentage Error (MPE) of the model predictions which is given by:
\begin{linenomath}
\begin{align}
&MPE=\frac{1}{T}\sum_{t=1}^{T}\frac{1}{|\mathcal{N}_t|}\sum_{i\in\mathcal{N}_t}\frac{y^i_{t}-\hat{y}^i_{t}}{y^i_{t}},\; \; \mathcal{N}_t = \{i: 1 \leq i \leq N, \; y^i_{t}>0.1\}
\end{align}
\end{linenomath}
The positive value of MPE indicates the underestimation of the TNC demand (i.e. the real demand is larger than the predicted demand), whereas the negative value of MPE indicates the overestimation of the TNC demand. The magnitude of a positive percentage error in region $i$ at time $t$ can be thought of the chance of an individual in region $i$ at time $t$ who had the TNC demand but failed to receive the service, if the TNC service was exactly allocated based on the TNC demand estimation. Therefore, it is important to make sure that the MPE is not systematically different between the disadvantaged and privileged communities. This concept is connected to one important notion of algorithmic fairness -- equality of odds, which states that a predictor $\hat{Y}$ satisfies equalized of odds with respect to protected attribute Z and outcome Y, if $\hat{Y}$ and Z are independent conditional on Y \cite{hardt2016equality}.\\

\noindent We propose the MPE gap as a fairness metric, which measures the difference of MPE between two groups (e.g. the black communities and the non-black communities). The metric is defined as:
\begin{linenomath}
\begin{equation}
\begin{split}
  MPE\:Gap={} & \frac{1}{T}\sum_{t=1}^{T}\frac{1}{|\mathcal{N}_{t,z_0}|}\sum_{i\in\mathcal{N}_{t,z_0}}\frac{y^i_{t}-\hat{y}^i_{t}}{y^i_{t}}-\frac{1}{T}\sum_{t=1}^{T}\frac{1}{|\mathcal{N}_{t,z_1}|}\sum_{i\in\mathcal{N}_{t,z_1}}\frac{y^i_{t}-\hat{y}^i_{t}}{y^i_{t}} \\
        & s.t. \; \; \mathcal{N}_{t,z_0} = \{i: 1 \leq i \leq N, \; y^i_{t}>0.1, \; i \in\mathcal{Z}_0\}, \; \;\mathcal{N}_{t,z_1} = \{i: 1 \leq i \leq N, \; y^i_{t}>0.1, \; i \in\mathcal{Z}_1\}
\end{split}
\end{equation}
\end{linenomath}
\noindent where ${Z}_0$ denotes the minority group and ${Z}_1$ denotes the majority group. Therefore, $i \in\mathcal{Z}_0$ represents the set of regions that are within the minority group, and $i \in\mathcal{Z}_1$ represents the set of regions that are within the majority group (i.e. not within the minority group). For example, if the sensitive variable of interest is ethnicity and ${Z}_0$ is used to represent the black-dominated communities, then ${Z}_1$ represents the non-black communities. In this case, $i \in\mathcal{Z}_0$ and $i \in\mathcal{Z}_1$ denote regions that belong to the black communities and those that belong to the non-black communities, respectively.\\

\noindent To achieve a fair prediction, we want the absolute value of MPE gap to be as close to zero as possible. A positive value of MPE gap indicates that we are underestimating the TNC demand for the minority group compared with the majority group, whereas a negative value of MPE gap suggests a relative underestimation of the demand for the majority group.


\subsection{4.6 De-biasing objective function}
\label{sec:obj_func}
To jointly train for accuracy and fairness, we use a loss function that is a weighted sum of an accuracy loss and a fairness loss defined as below:
\begin{linenomath}
\begin{align}
\label{equ:obj_main}
L={} & L_{accuracy}+\gamma{}L_{fairness}
\end{align}
\end{linenomath}
The accuracy loss is aimed at reducing both MAE and MAPE:
\begin{linenomath}
\begin{align}
L_{accuracy}={} & \sum_{t=1}^{T}\sum_{i=1}^{N} (y^i_{t}-\hat{y}^i_{t})^2+\lambda \sum_{t=1}^{T}\sum_{i\in\mathcal{N}_t}(\frac{y^i_{t}-\hat{y}^i_{t}}{y^i_{t}})^2, \; \;s.t.\; \; \mathcal{N}_t = \{i: 1 \leq i \leq N, \; y^i_{t}>0.1\}
\end{align}
\end{linenomath}
where $y^i_{t}$ and $\hat{y}^i_{t}$ are the real and predicted travel demands at time interval $t$ in region $i$. $T$ represents the total number of time intervals. $N$ represents the total number of regions. $\mathcal{N}_t$ denotes the set of regions with $y^i_{t}>0.1$. $\lambda$ is a regularization parameter balancing the MAE and MAPE tradeoff. In this study, we fix $\lambda$ to be 10 since the magnitude of MAE is roughly ten times that of MAPE.\\

\noindent The fairness loss is proposed as the following:
\begin{linenomath}
\begin{align}
\label{equ:L_fairness}
L_{fairness}={} & |\sum_{t=1}^{T}\sum_{i\in\mathcal{N}_t}\tilde{z}^i*\frac{y^i_{t}-\hat{y}^i_{t}}{y^i_{t}}|, \; \;s.t.\; \; \tilde{z}^i=\frac{z^i-\Bar{z}}{\sigma_z}, \; \; \mathcal{N}_t = \{i: 1 \leq i \leq N, \; y^i_{t}>0.1\}
\end{align}
\end{linenomath}
where $z^i$ denotes the value of the sensitive attribute (e.g. the proportion of black population) for region $i$. $\tilde{z}^i$ is the normalized $z^i$ with $\Bar{z}$ and $\sigma_z$ respectively representing the mean and standard deviation of $z^i$ across all regions.\\

\noindent $L_{fairness}$ measures the linear relationship between the sensitive attribute $z$ and MPE across time and space. To be specific, $\tilde{z}^i*\frac{y^i_{t}-\hat{y}^i_{t}}{y^i_{t}}$ measures the joint deviations of $\tilde{z}^i$ and $\frac{y^i_{t}-\hat{y}^i_{t}}{y^i_{t}}$ from zero. Therefore, $L_{fairness}$ indicates the covariance between $z$ and MPE in the prediction, which we want to penalize in our training process.

\section{5. Experiments}
\label{sec:experiment}
\subsection{5.1 Data Description}
\begin{figure}
\centering
\subfloat[Average hourly TNC demand]{\includegraphics[width=2in]{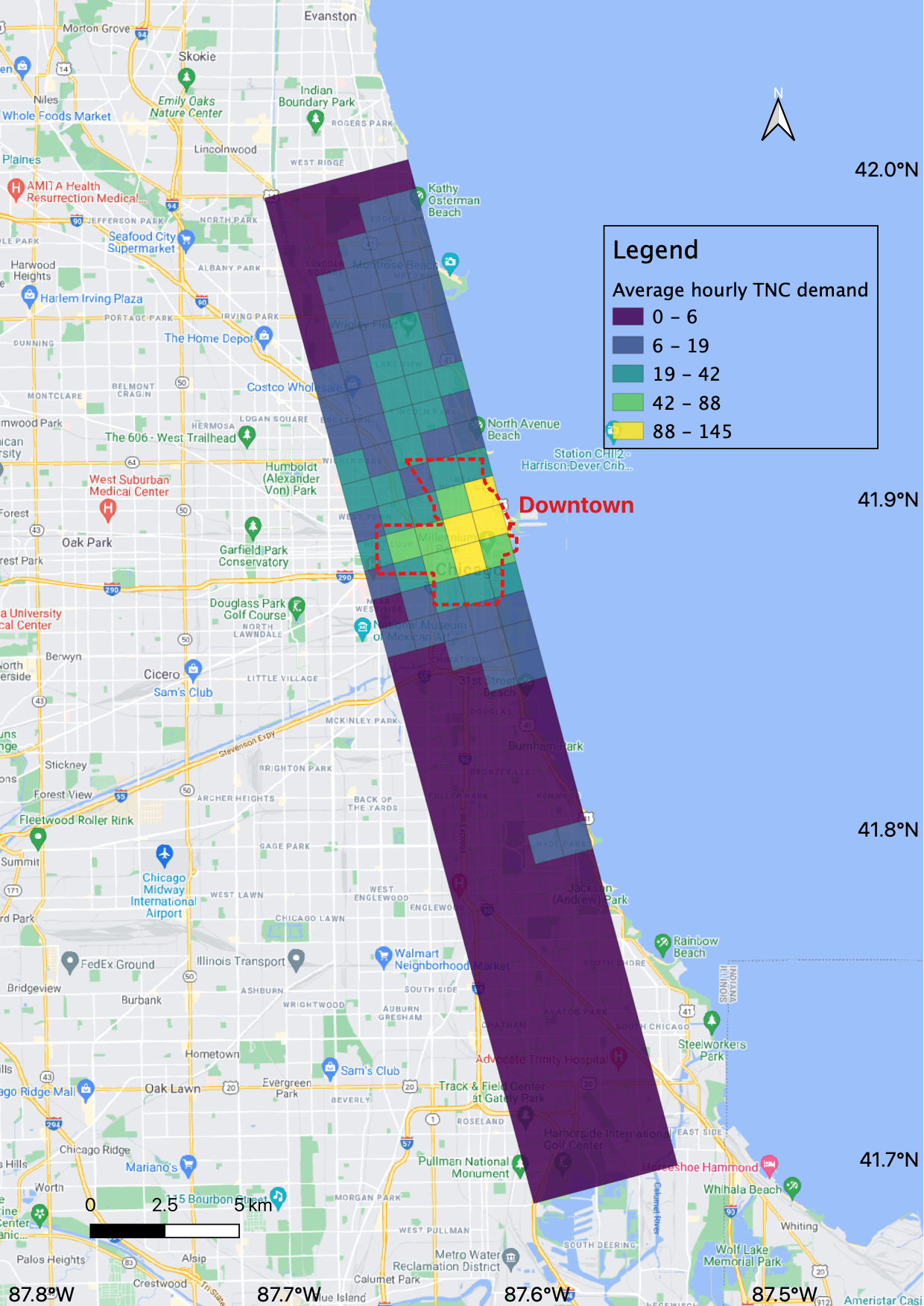}\label{fig:desc_tnc}} 
\quad
\subfloat[Percentage of black population]{\label{fig:desc_black}\includegraphics[width =2in]{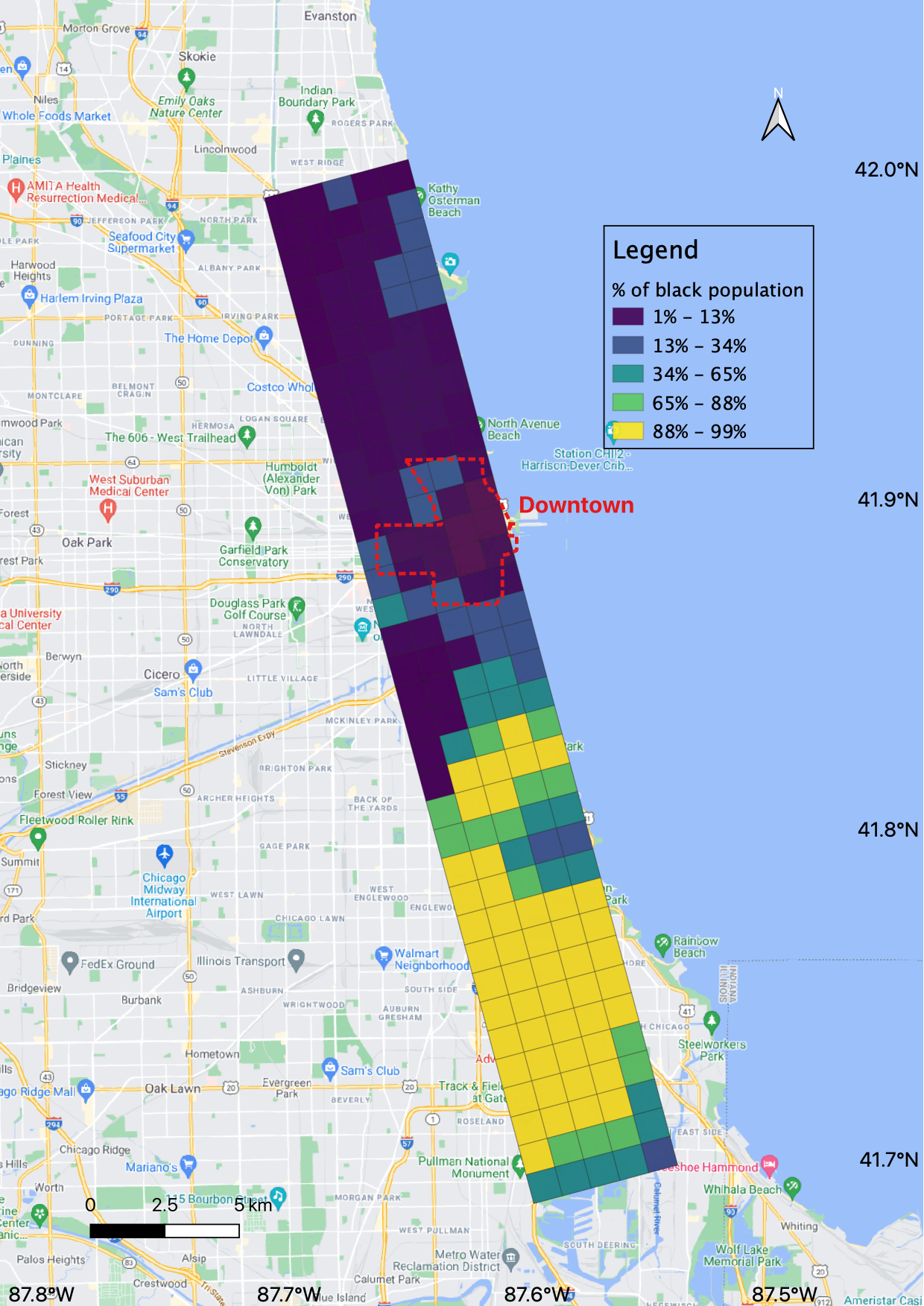}}
\quad
\subfloat[Percentage of low-income population]{\label{fig:desc_poverty}\includegraphics[width =2in]{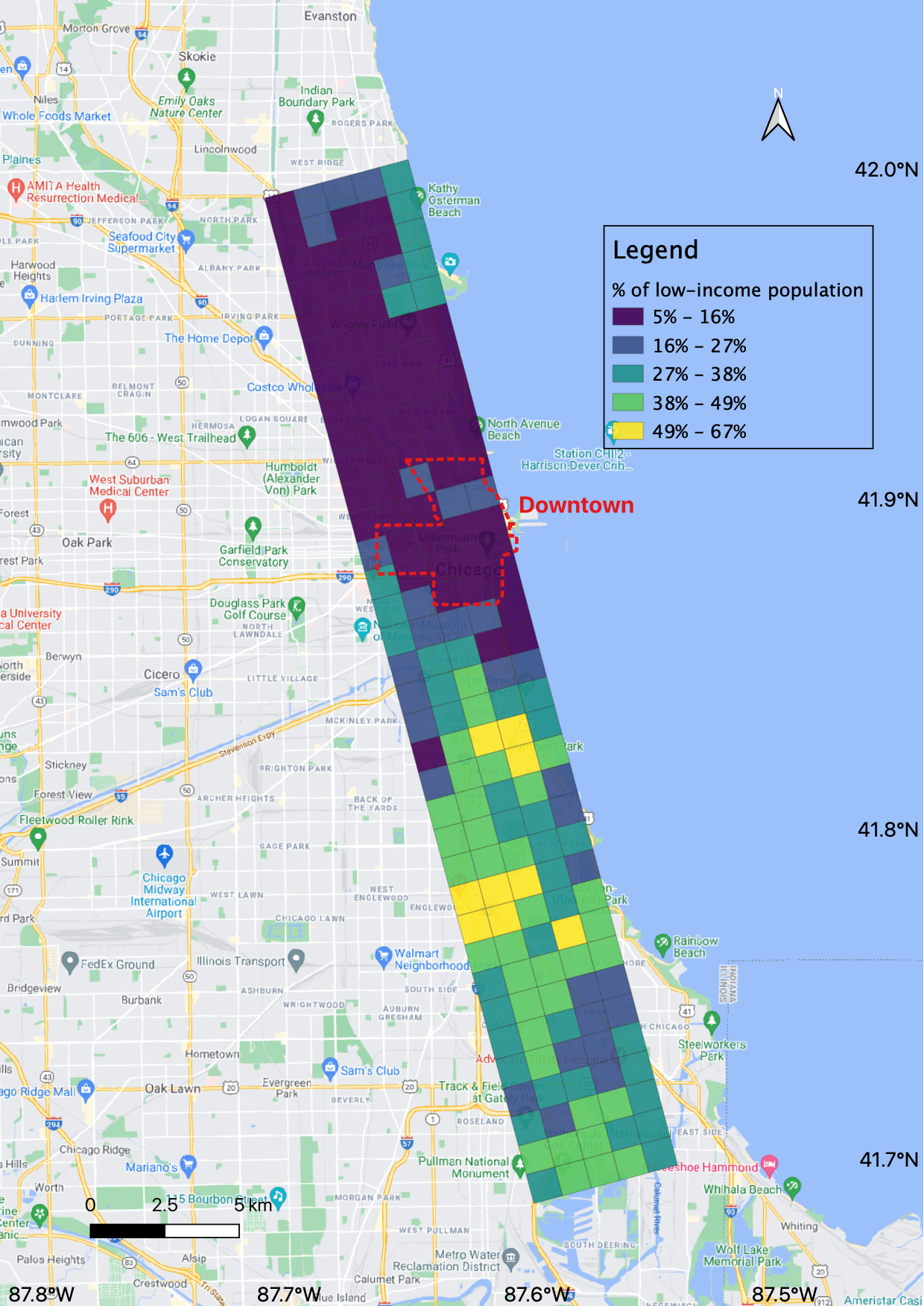}}
\caption{Distributions of TNC demand, black population and low-income population in the study area} 
\label{fig_desc} 
\end{figure}

The dataset utilized in this paper is a large-scale TNC trip record dataset collected from Chicago Data Portal \cite{portal2020transportation} during a 14-month period between November 1st, 2018 to December 23rd, 2019. 
The trip records that started from 6 AM to 10 PM are included. We partition the city of Chicago into $1km \times 1km$ grids, and use totally $35 \times 5$ grids for analysis as shown in Figure \ref{fig_desc}. The hourly TNC demand in a region is represented by the number of trips starting from that region in a 1-hour time interval. The weather data is collected from the website of National Centers for Environmental Information \cite{national2021climate}. The socio-demographic variables including the percentages of black population and the percentage of low-income population are extracted from the 2019 American Community Survey(ACS) 5-year estimates \cite{us20192015}.\\

\noindent Figure \ref{fig_desc} illustrates the distributions of average hourly TNC demand in the study period, the percentage of black population and the percentage of low-income population in the study area. From Figure \ref{fig:desc_tnc}, we can see that the spatial distribution of the TNC demand is highly uneven, as the downtown area takes up the majority of the TNC demand. 
In terms of ethnicity, Figure \ref{fig:desc_black} reveals a bimodal distribution of African-American population, with the majority of the northern area having African-American population below 13\% and the majority of the southern area having African-American population above 88\%. We define population with 2019 household income lower than \$25,000 as low-income, and Figure \ref{fig:desc_poverty} shows that the low-income population is also mainly clustered in the south side of the study area. In this study, we define grids with over 50\% of black population as the black communities, and the rest as the non-black communities, which gives us 73 black communities and 102 non-black communities. Regarding income, we defined grids with more than 25\% of low-income population as the low-income communities, and the rest as the high-income communities, resulting in totally 90 low-income communities and 85 high-income communities. In both cases, the numbers of disadvantaged and privileged communities are roughly balanced.\\

\begin{figure}
\centering
\subfloat[Disadvantaged regions]{\label{fig:true_min}\includegraphics[width=3in]{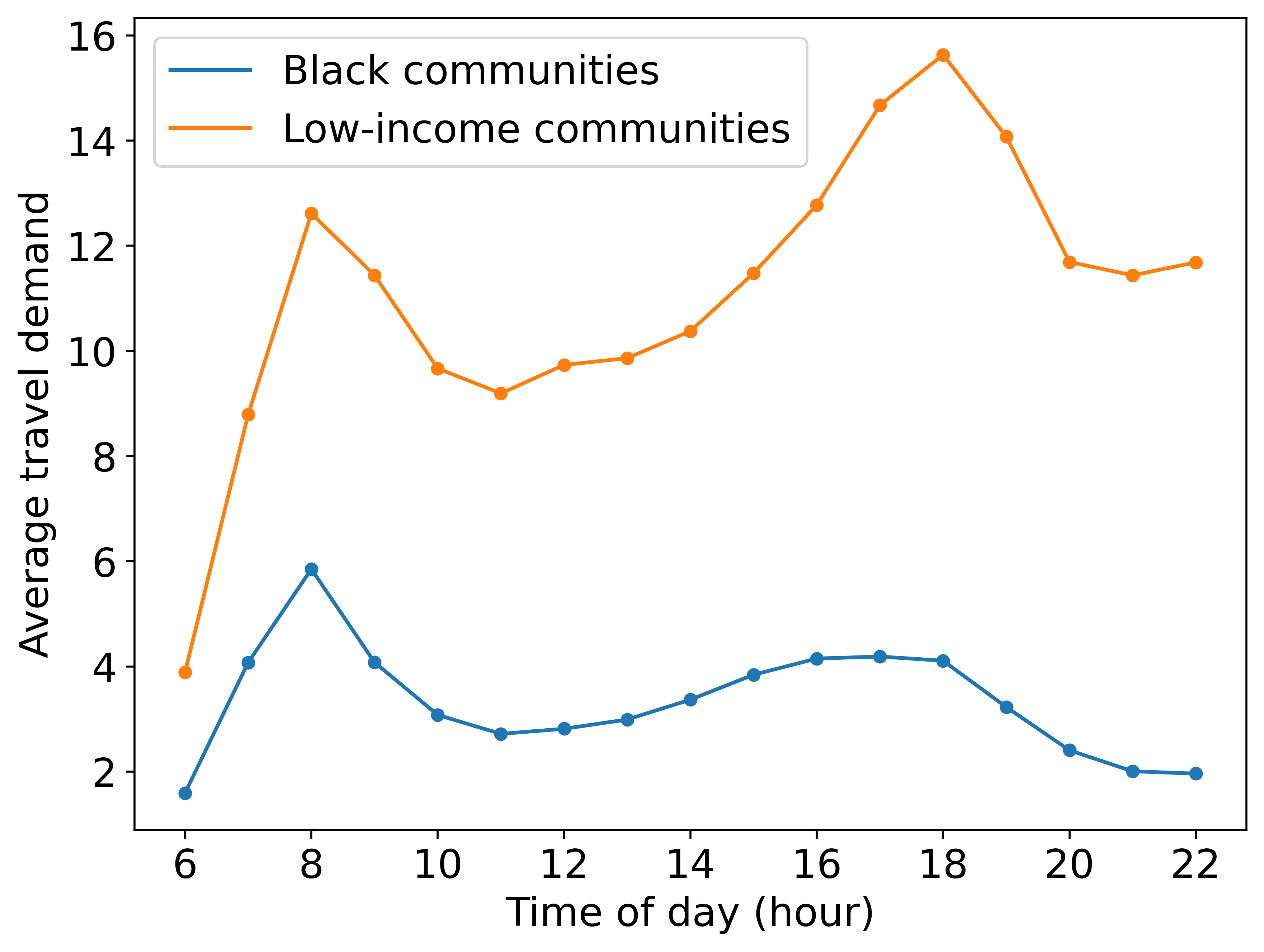}} 
\qquad
\subfloat[Privileged regions]{\label{fig:true_maj}\includegraphics[width =3in]{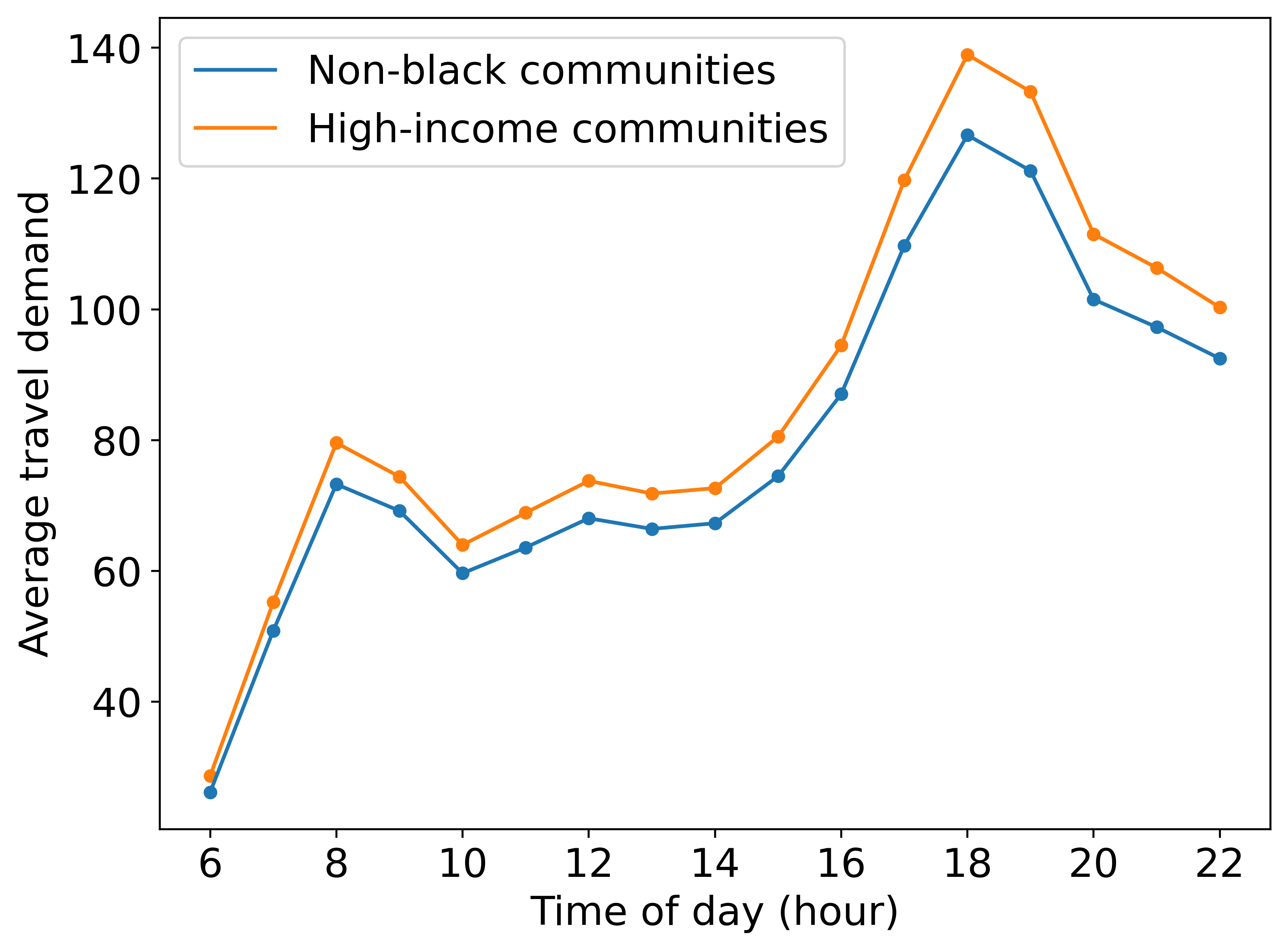}}
\caption{Average TNC travel demand by time of day} 
\label{fig_true_ts} 
\end{figure}

\noindent Figure \ref{fig_true_ts} illustrates the average TNC travel demand by time of day, separated by the disadvantaged (black/low-income) and privileged (non-black/high-income) communities. The travel demand in the privileged regions are much larger than that in the disadvantaged regions, therefore the y-axis scales are different in Figure \ref{fig_true_ts}(a) and Figure \ref{fig_true_ts}(b). The privileged regions and the low-income communities have two peak periods: 7 AM - 10 AM and 5 PM - 8 PM, whereas the black communities only has the morning peak.\\

\noindent In the experiment, we apply a 70-30 training-testing split. The data from November 1st, 2018 to July 23rd, 2019 (265 days) is used for training, the data from July 24th, 2019 to August 21st, 2019 (29 days) is used for validation and the data from August 22nd, 2019 to December 23rd, 2019 (124 days) is used for testing. In the training, validation and testing processes, we use the TNC demand in the previous 6 hours to predict the TNC demand in the next hour (i.e. the look-back window is 6 hours). Before training the models, the collected data is normalized by z-score process to facilitate training. We later denormalize the prediction to get the actual demand values, and reset the negative values to zeros since the demand values cannot be negative. 

\subsection{5.2 Model Comparison}
\label{sec:model_comparison}
To explore the advantage of our model SA-Net, we compare it against several other benchmark models, which are listed as follows:
\begin{itemize}
    \item \textbf{Historical Average (HA)}: HA predicts the TNC demand by averaging the historical demand which is in the same relative time interval (i.e. the same time of day and the same day of week) in the training set. For instance, the TNC demand in Monday 10 AM -11 AM is predicted as the average TNC demand of all past Monday's at 10 AM -11 AM in the training set.
    \item \textbf{Moving Average (MA)}: MA predicts the TNC demand by averaging the demand in the same relative time interval of several nearest historical values. We use the average of 6 previous TNC demand in grid $(i,j)$ to predict the demand in grid $(i,j)$.
    \item \textbf{Autoregressive Integrated Moving Average Model (ARIMA)}: ARIMA is commonly used for forecasting time-series data \cite{box1970distribution}, and has been widely applied in traffic prediction problems\cite{van1996combining,williams2003modeling}. In this work, to predict the TNC demand in grid $(i,j)$, the inputs to ARIMA were 6 previous demand in the same relative time interval in grid $(i,j)$.
    \item \textbf{LSTM Net}: The LSTM Net processes the TNC demand in each grid separately. The hyperparameters and the structures of the LSTM Net and the SA-Net are the same. The only difference is that while we use a stack of SAC-LSTM to process the spatial-temporal data as shown in Figure \ref{fig:model}, the LSTM Net uses the LSTM modules to processes the TNC demand data and does not capture spatial dependencies. 
    \item \textbf{LSTM + Social Net}: The LSTM + Social Net adds a socio-demographic feature map to the LSTM Net to facilitate predictions. The feature map is constructed as a linear combination of different socio-demographic variables as shown in Figure \ref{fig:feature_map}, and is fused with other parts of the network in the last model layer following Equation \ref{eq_fuse}.
    \item \textbf{Conv-LSTM Net}: The Conv-LSTM Net is a fusion convolutional LSTM specified in \cite{ke2017short}. The hyperparamters and the structure of the Conv-LSTM Net are the same with the SA-Net and the LSTM Net. The difference is that the Conv-LSTM Net uses the traditional Conv-LSTM modules instead of the SAC-LSTM modules in Figure \ref{fig:model} to process the spatial-temporal TNC demand data. 
    \item \textbf{Conv-LSTM + Social Net}: Similar to the LSTM + Social Net, the Conv-LSTM + Social Net adds a socio-demographic feature map to the Conv-LSTM Net to facilitate predictions. 

\end{itemize}

\subsection{5.3 Experiment setup}
When training Conv-LSTM Net and SA-Net, we use kernels with size of $3 \times 3$. Each Conv-LSTM cell and each SAC-LSTM cell consists of 64 filters/channels to capture the spatial information. The experiments are implemented in Pytorch using the mini-batch stochastic gradient descent method with a batch size of 64 and a step size of 0.001 in each training. The
model that produces the lowest prediction loss on the validation set among the 300 epochs is chosen. For both Conv-LSTM Net and SA-Net, we train the model with the number of layers being 1, 2 and 3 and choose the one that produces the lowest prediction loss. The optimal model later performs prediction over the test data. We run the training procedure 3 times and report the average prediction results on the test set.

\subsection{5.4 Results}
We compare our proposed algorithms (SA-Net with bias-mitigation regularization) with baseline models along two dimensions: accuracy and fairness, and show that our algorithm achieves better results regarding both accuracy and fairness. The better prediction accuracy is demonstrated by lower MAE and MAPE. The better prediction fairness is shown in terms of two aspects: first, we will show that our model reduces MAE for the disadvantaged groups to a greater extent than the privileged groups compared with Conv-LSTM Net; second, 
our results show that the proposed bias mitigation strategy can reduce the MPE gap between disadvantaged and privileged groups while not harming the overall prediction accuracy, thus achieving a fair prediction.

\subsubsection{5.4.1 Prediction accuracy}
The spatial-temporal deep learning algorithms (i.e. Conv-LSTM Net, Conv-LSTM + Social Net and SA-Net) outperform the classical statistical models (i.e. HA, MA, ARIMA), and our proposed SA-Net model produces the smallest overall MAE and MAPE among all models on the test set. Table \ref{tab:result_MAE} reports the overall MAE and MAPE, as well as the MAE and MAPE for the black, non-black, low-income and high-income communities. Regarding the overall MAE and MAPE, Conv-LSTM Nets and SA-Net significantly outperform other models, indicating the predictive power of these models which capture both temporal and spatial dependencies. Conv-LSTM + Social Net performs slightly worse than Conv-LSTM Net, probably because of overfitting of the model. This result suggests that simply adding the socio-demographic variables as predictors does not improve the model performance. We now compare the results between the two best-performing models regarding the overall MAE and MAPE: Conv-LSTM Net and SA-Net.\\

\noindent Comparing Conv-LSTM Net and SA-Net, we find that SA-Net reduces MAE for both the black and non-black communities. It also reduces MAE for both the low-income and the high-income communities. These results indicate that by incorporating the socio-demographic information, SA-Net benefits both disadvantaged and privileged groups. In addition, the disadvantaged communities experience a larger decrease in MAE with SA-Net compared with Conv-LSTM Net. From Conv-LSTM Net to SA-Net, the reductions in MAE for the black and non-black populations are respectively 0.12 and 0.05, and the reductions in MAE for the low-income and high-income populations are 0.10 and 0.03, respectively. This result shows that our proposed SA-Net can not only improve the overall model performance, but can also promote fairness in predictions by increasing more prediction accuracy for the disadvantaged populations while not harming the performance of the privileged populations.


\begin{table}[H]\centering 
  \caption{Accuracy comparisons among different models}
  \label{tab:result_MAE} 
\resizebox{0.9\textwidth}{!}{%
\begin{tabular}{@{}lllllll@{}}
\\[-1.8ex]\hline 
\hline \\[-1.8ex] 
                             & MAE  & MAPE & Black & Non-black & Low-income& High-income \\
 &  &  & (MAE) & (MAE) & (MAE) & (MAE) \\\midrule
HA                     & 8.00 & 0.64 & 1.55        & 12.61           & 6.53             & 12.11             \\
MA                     & 7.93 & 0.55 & 1.41        & 12.60           & 6.49             & 11.96             \\
ARIMA                  & 7.16 & 0.54 & 1.36        & 11.31           & 5.93             & 10.60             \\
LSTM Net               & 8.81 & 0.49 & 2.08        & 13.62           & 6.97             & 13.97             \\
LSTM + Social Net      & 8.85 & 0.51 & 2.09        & 13.69           & 7.00             & 14.03             \\
Conv-LSTM Net          & 6.36 & 0.43 & 1.81        & 9.61            & 5.21             & 9.58              \\
Conv-LSTM + Social Net & 6.42 & 0.44 & 1.82        & 9.71            & 5.25             & 9.69              \\
SA-Net                 & 6.28 & 0.41 & 1.69        & 9.56            & 5.11             & 9.55              \\
\hline 
\hline \\[-1.8ex] 
\multicolumn{7}{l}{\textit{Note:} for the deep learning models, we report the results when the models are trained with $\gamma = 0$}\\  
\end{tabular}%
} 
\end{table}

\begin{figure}[!h]
\centering
\subfloat[MAE: black]{\label{fig:MAE_min_black}\includegraphics[width=2in]{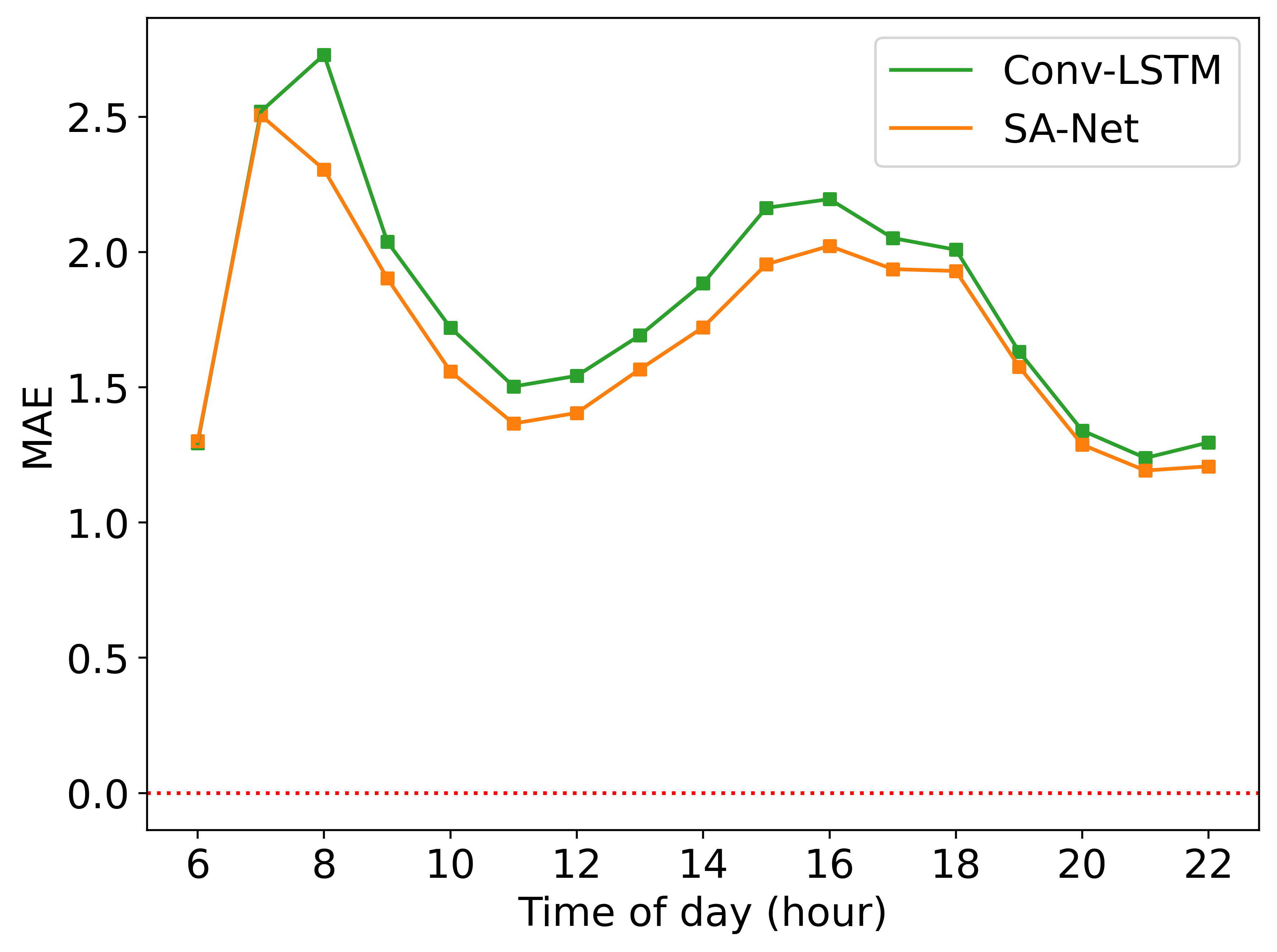}} 
\quad
\subfloat[MAPE: black]{\label{fig:MAPE_min_black}\includegraphics[width =2in]{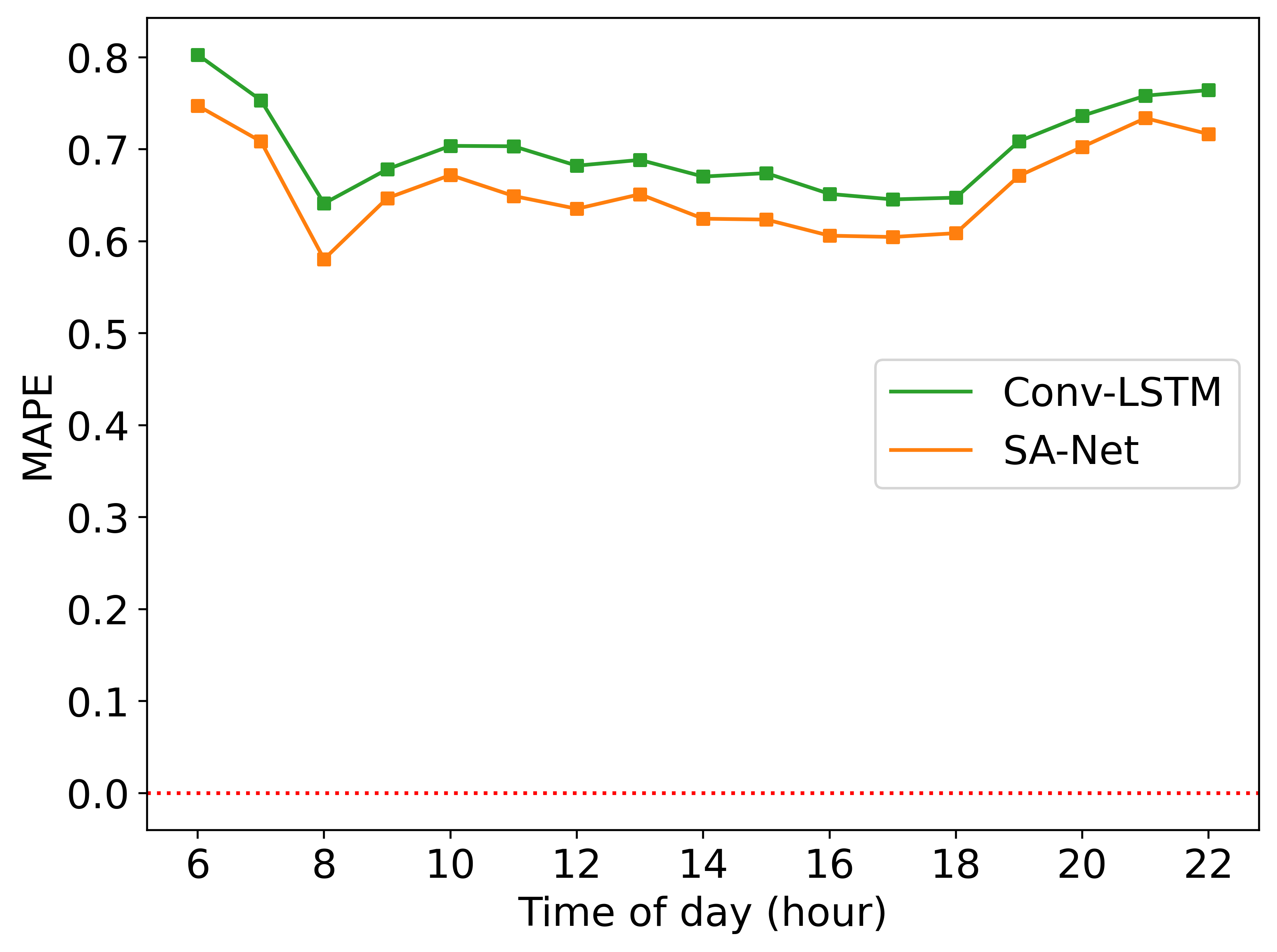}}
\quad
\subfloat[MPE: black]{\label{fig:MPE_min_black}\includegraphics[width =2in]{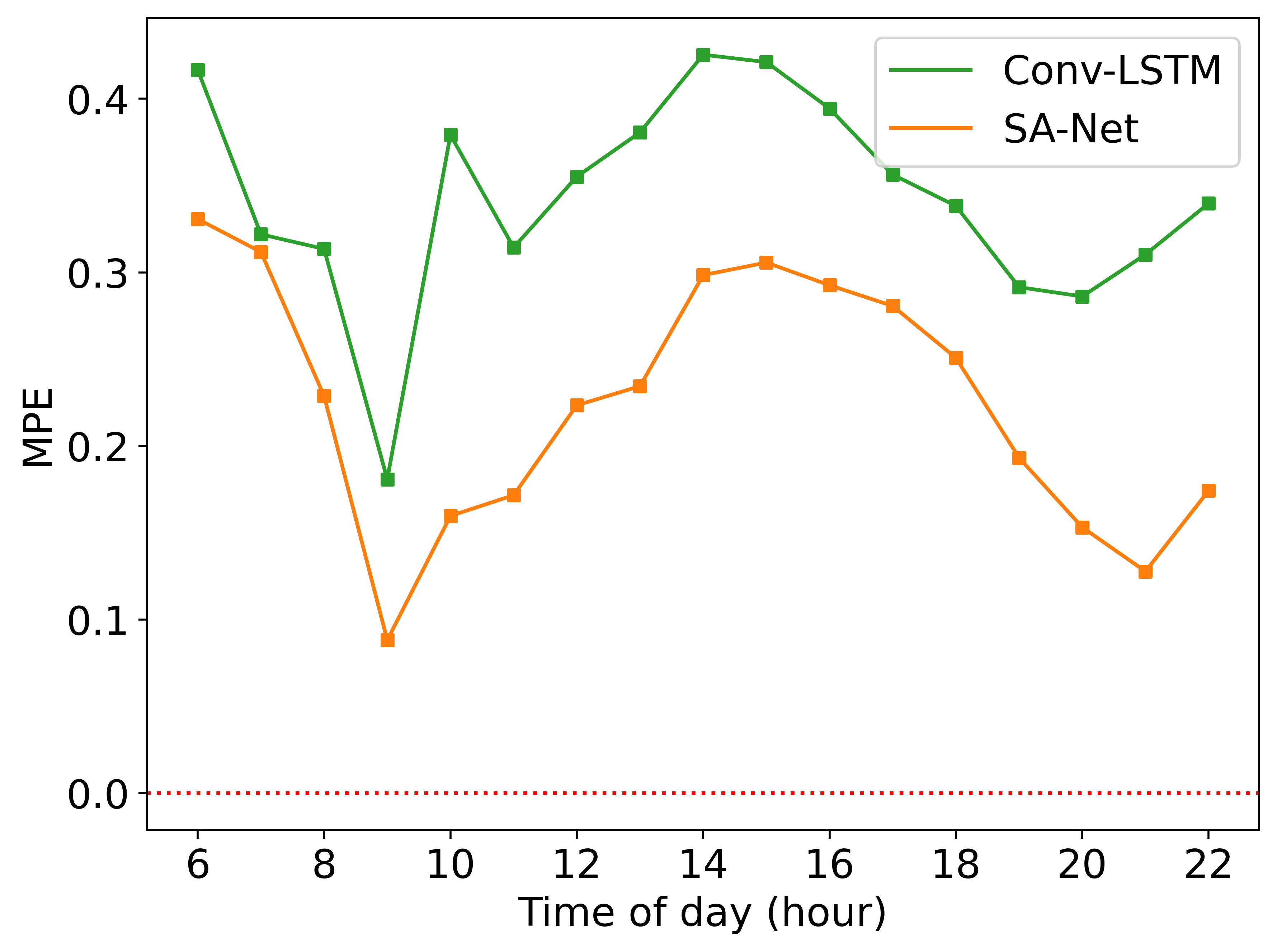}}\par\medskip

\subfloat[MAE: non-black]{\label{fig:MAE_maj_black}\includegraphics[width=2in]{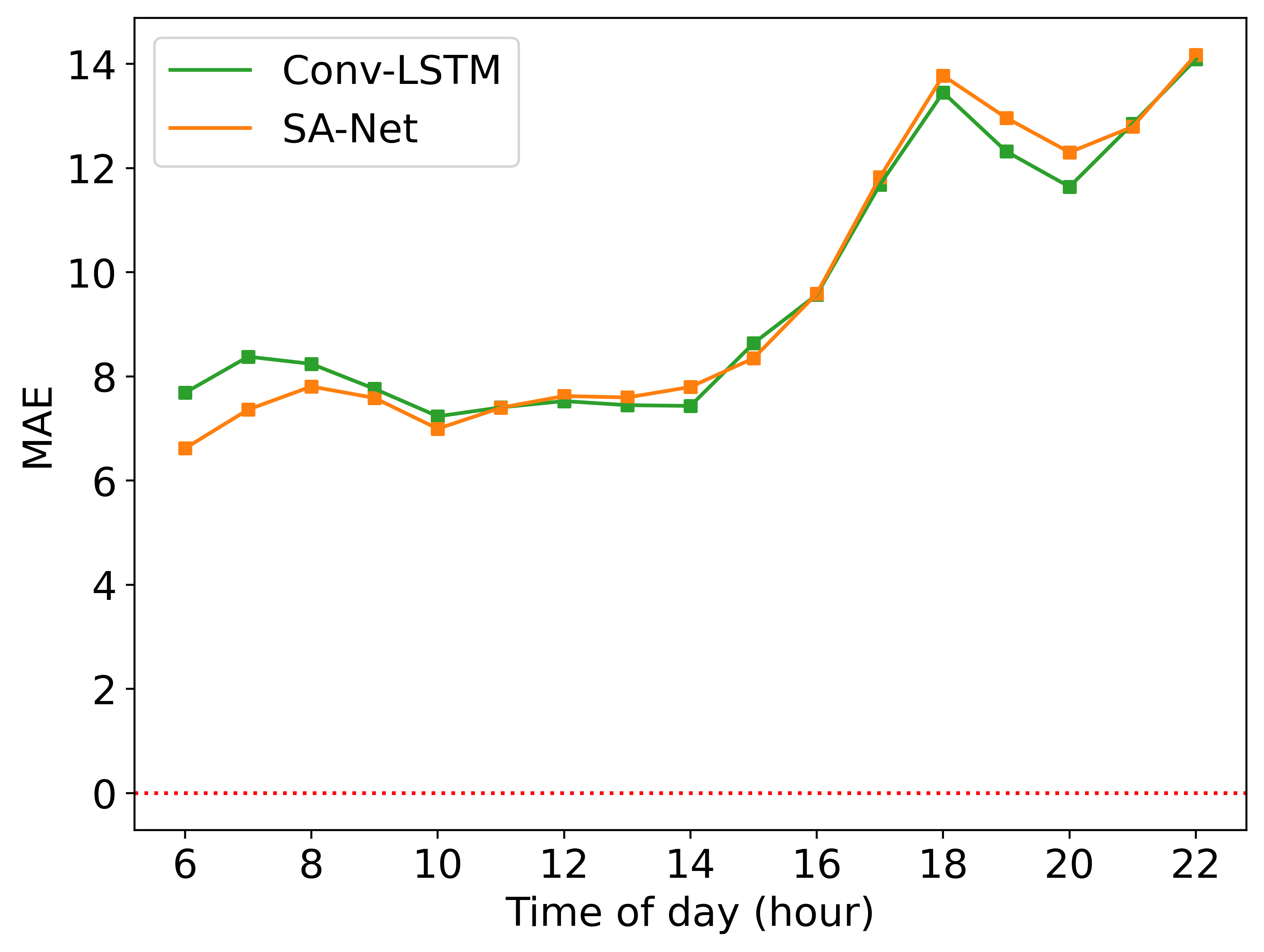}} 
\quad
\subfloat[MAPE: non-black]{\label{fig:MAPE_maj_black}\includegraphics[width =2in]{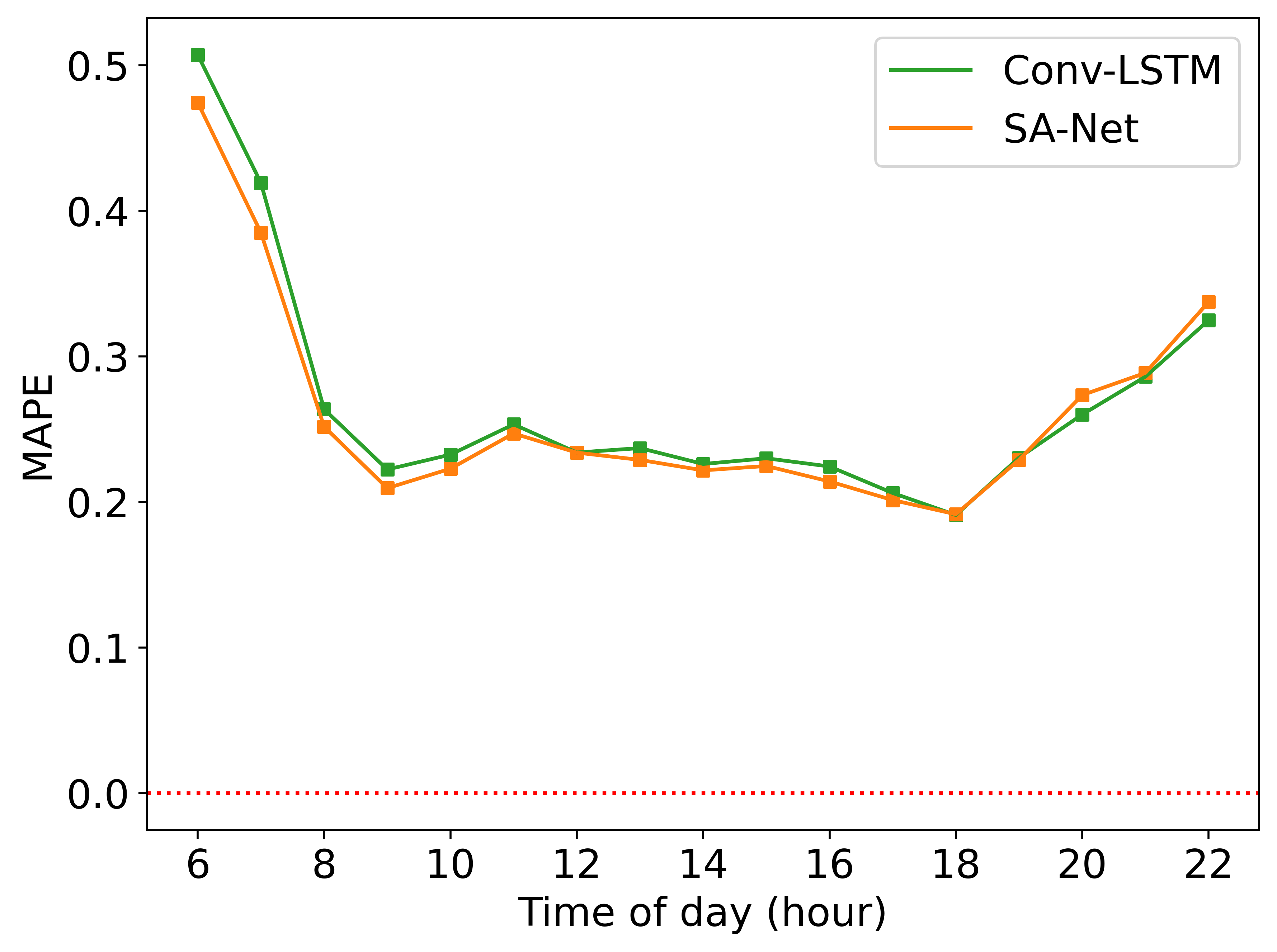}}
\quad
\subfloat[MPE: non-black]{\label{fig:MPE_maj_black}\includegraphics[width =2in]{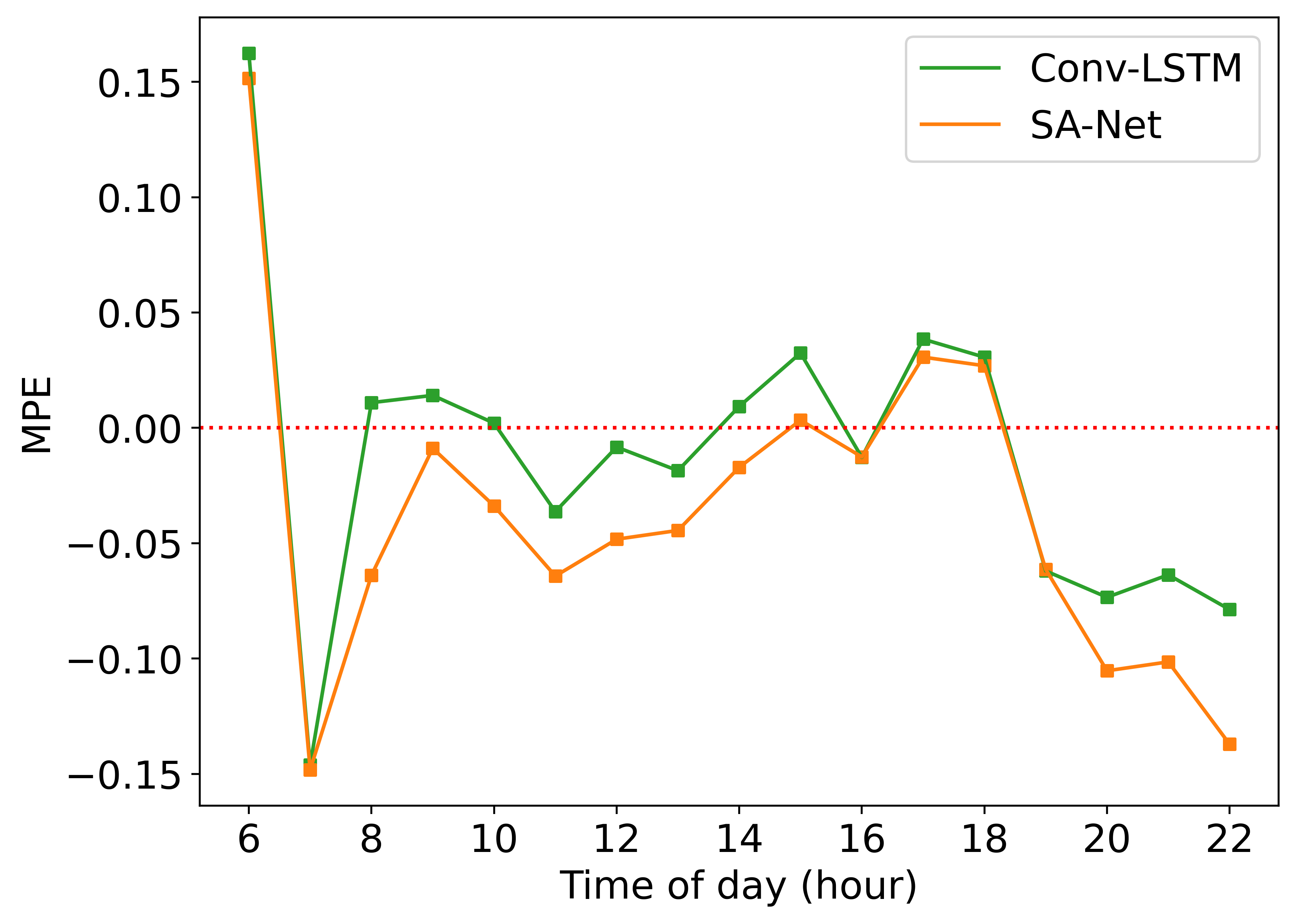}}
\caption{Performance measures by model and time of day} 
\label{fig_chart_ts1} 
\end{figure}

\noindent Next, SA-Net improves prediction accuracy for the black communities at all times of day compared with Conv-LSTM Net. We examine the model performance for Conv-LSTM Net and SA-Net across different times of day in Figure \ref{fig_chart_ts1}. The upper row of Figure \ref{fig_chart_ts1} shows the predictive results for the black communities, whereas the bottom row of Figure \ref{fig_chart_ts1} shows the results for the non-black communities. Figure \ref{fig:MAE_min_black} and \ref{fig:MAPE_min_black} show that SA-Net produces smaller MAE and MAPE for all times of day than Conv-LSTM Net for the black communities. Figure \ref{fig:MAE_maj_black} and \ref{fig:MAPE_maj_black} show that compared with Conv-LSTM Net, SA-Net produces smaller MAE and MAPE in the morning period (6 AM - 10 AM), but produces slightly larger MAE and MAPE in the evening peak period (6 PM - 8 PM) for the non-black communities.\\

\noindent Figure \ref{fig:MPE_min_black} shows the MPE for the black communities. The MPEs are consistently positive, indicating that both Conv-LSTM Net and SA-Net underpredict the travel demand for the black communities at different times of day. However, SA-Net consistently gives smaller MPEs than Conv-LSTM Net, showing that the former model reduces the magnitude of the underprediction of the black communities' travel demand. The MPEs for the non-black communities with Conv-LSTM Net and SA-Net are more similar at different times of day as shown in Figure \ref{fig:MPE_maj_black}

\FloatBarrier
\FloatBarrier

\noindent Nevertheless, it is worth noting that the MAEs for the black communities generated by the deep learning models are always higher than those generated by the classical statistical approaches (i.e. HA, MA, ARIMA) as shown in Table \ref{tab:result_MAE}. These results may indicate that while the deep learning models significantly advances the overall prediction accuracy, this accuracy gain mainly comes from the improved model performance for the privileged communities, while the prediction accuracy for the disadvantaged communities could become worse. This highlights the necessity of minimizing the potential harms for the disadvantaged groups while scholars are increasingly in favor of the deep learning models over the traditional methods, and our approach contributes to this goal by bringing down the MAE for the disadvantaged communities in deep learning models. 


\subsubsection{5.4.2 Prediction fairness}
\begin{figure}
\centering
\subfloat[Conv-LSTM Net: race]{\label{fig:conv_black}\includegraphics[width=3in]{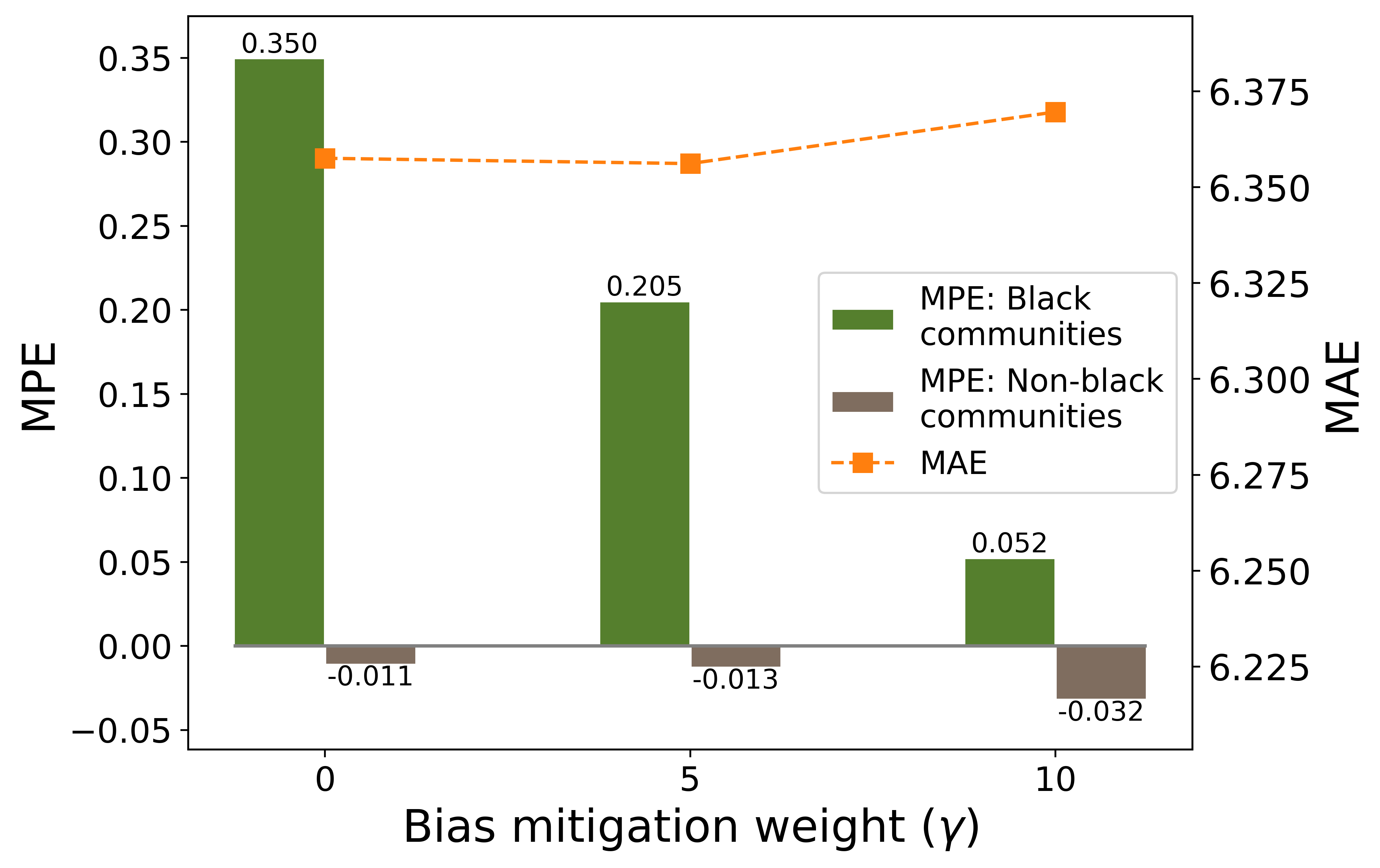}} 
\qquad
\subfloat[SA-Net: race]{\label{fig:pac_black}\includegraphics[width =3in]{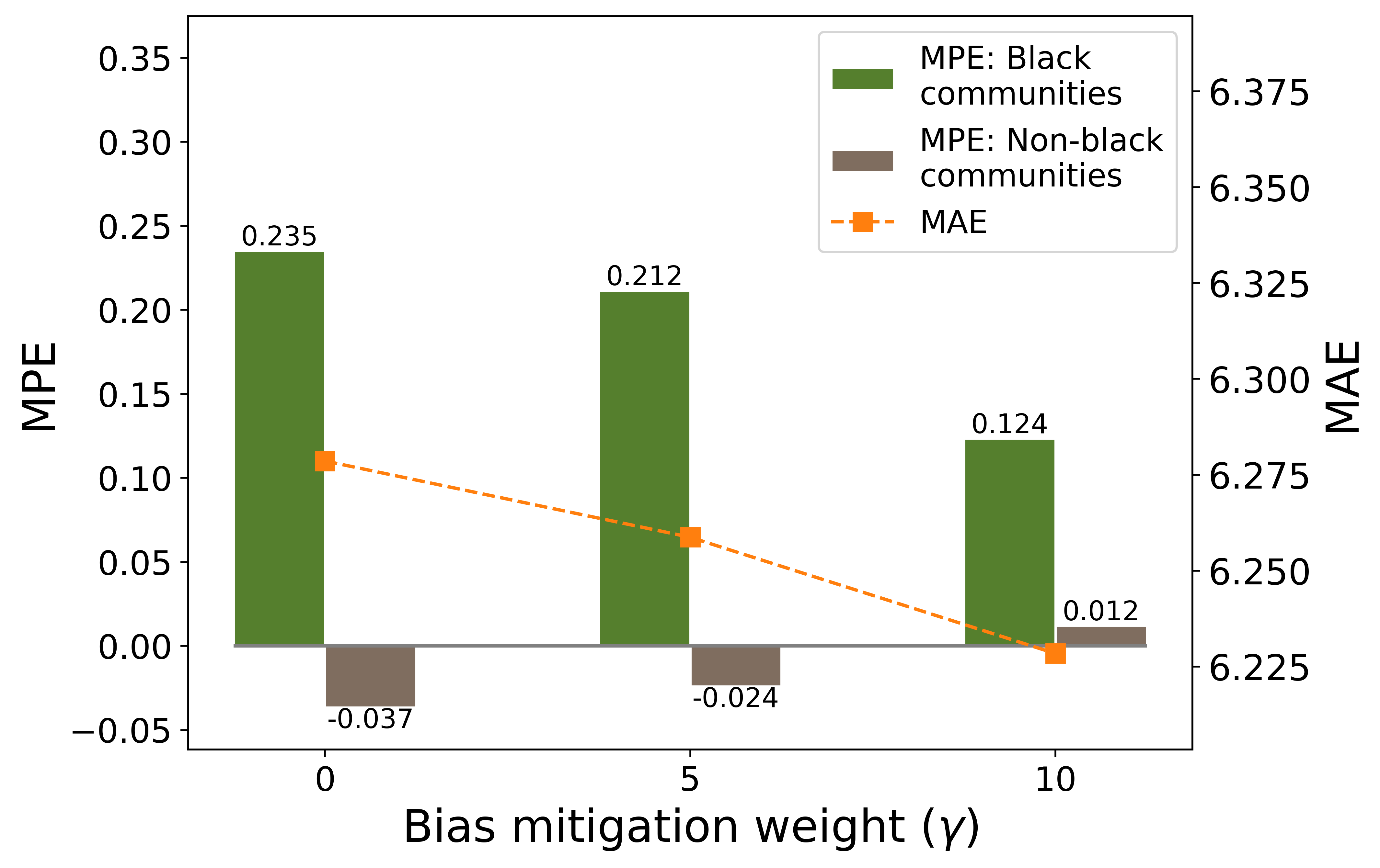}}\par\medskip

\subfloat[Conv-LSTM Net: income]{\label{fig:conv_income}\includegraphics[width=3in]{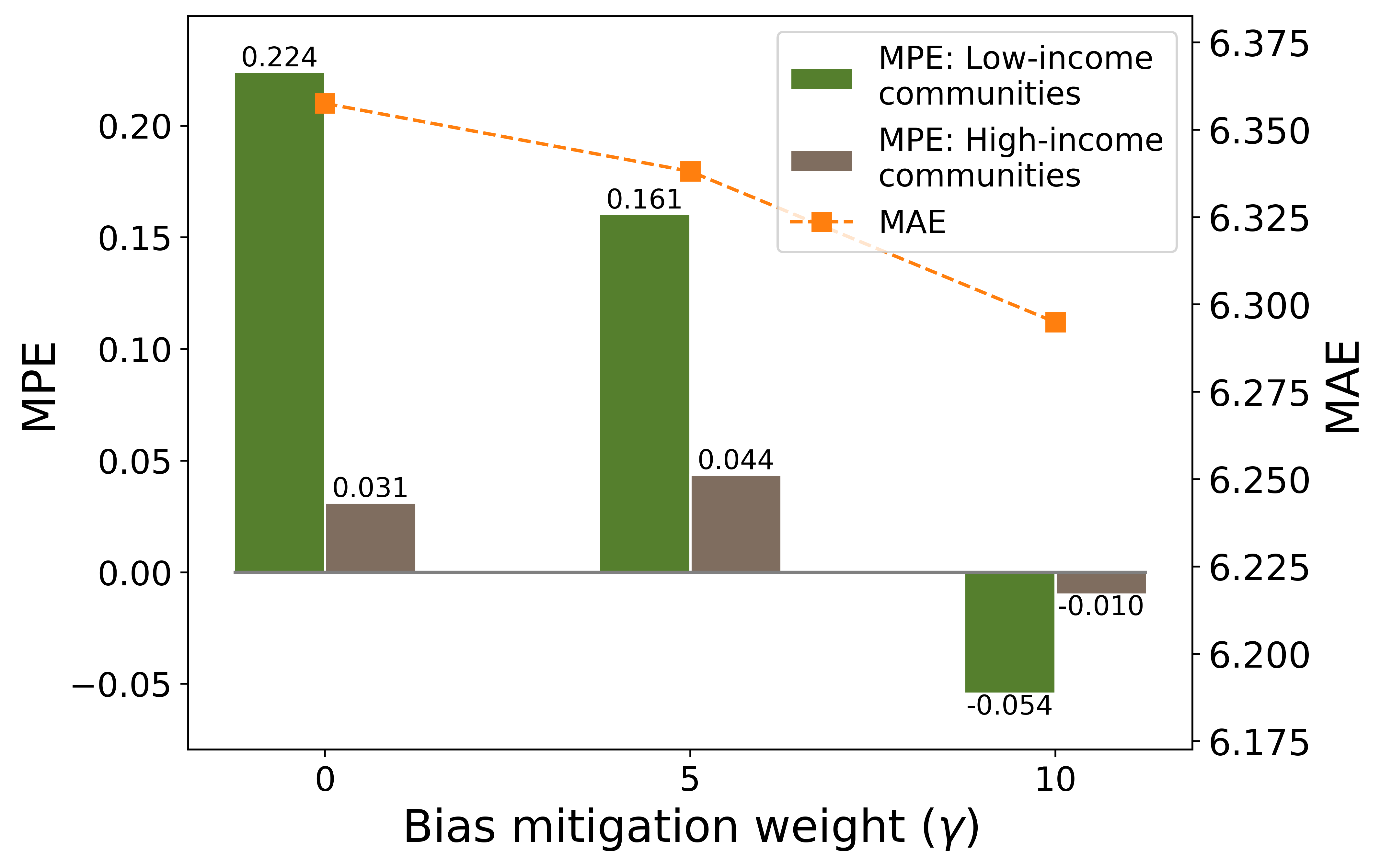}}
\qquad
\subfloat[SA-Net: income]{\label{fig:pac_income}\includegraphics[width =3in]{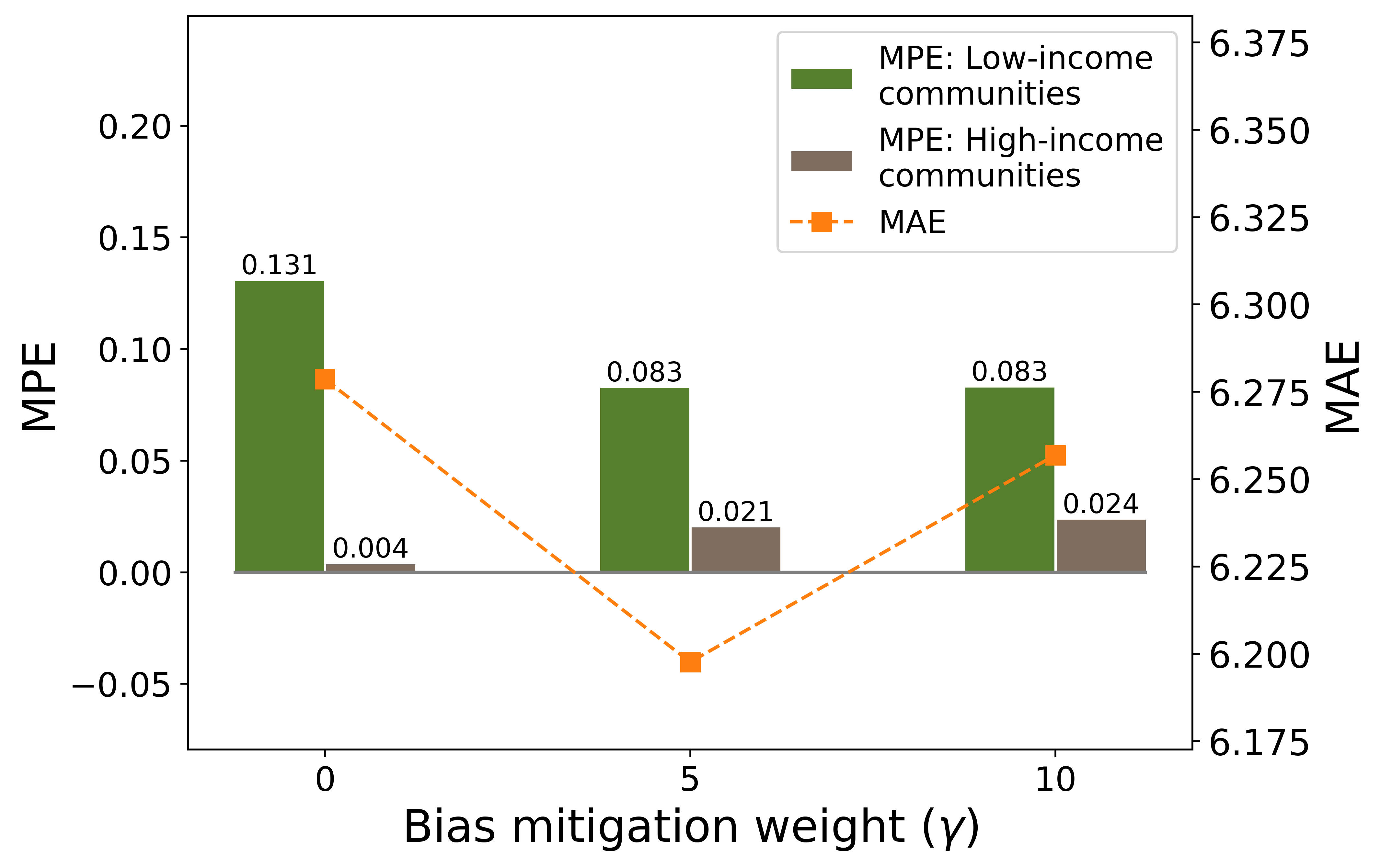}}
\caption{Performance measures by model and sensitive variable, corresponding to Table \ref{tab:result_bias_race} and Table \ref{tab:result_bias_income}} 
\label{fig_chart} 
\end{figure}

Having demonstrated the superiority of our proposed SA-Net over the benchmark models in terms of prediction accuracy, we now test the effectiveness of our bias mitigation strategy stated in Section \ref{sec:obj_func} for fairness improvement. First, we test the results when the sensitive attribute is race, namely when $z$ denotes the proportion of black population in Equation \ref{equ:L_fairness}. Table \ref{tab:result_bias_race} presents the results, and Figures \ref{fig:conv_black} and \ref{fig:pac_black} plot MPE for black and non-black groups as well as the overall MAE. Table \ref{tab:result_bias_race} shows that when there the de-biasing regularizer is not applied ($\gamma=0$), both Conv-LSTM Net and SA-Net produce large MPE gaps between black and non-black groups. Specifically, the MPE gap (race) with $\gamma=0$ is 0.361 for Conv-LSTM Net, whereas the MPE gap (race) with $\gamma=0$ is 0.272 for SA-Net. For both models, the large MPE gap comes from a large, positive MPE for the black group and a small, negative MPE for the non-black group. Note that the magnitude of a positive MPE indicates the degree of underestimation of the demand, since MPE represents the average gap of the actual and predicted demand weighted by the actual demand. Larger the MPE, higher the underestimation. Therefore, the large MPE gaps between black and non-black groups indicate that training models using the traditional objective function without bias mitigation leads to systematic underestimation for the black group compared with the non-black group.\\

\noindent Recognizing the prediction bias using only $L_{accuracy}$ in training, we adopt bias mitigation by increasing the bias mitigation weight $\gamma$ from 0 to 5 and 10. The results for ``MPE gap (race)" in Table \ref{tab:result_bias_race} show that for both models, as $\gamma$ increases, the MPE gap between black and non-black groups decreases, and this reduction in MPE gap mainly stems from the reduction in MPE for the black group. Specifically, when increasing $\gamma$ from 0 to 10, the MPE gap between the black and non-black groups drops from 0.361 to 0.084 for Conv-LSTM Net, and drops from 0.272 to 0.111 for SA-Net. It is also found that by mitigating the racial bias, the MPE gap between the low-income and high-income groups has also been reduced, probably because most low-income and black communities are clustered in the south side of Chicago (Figure \ref{fig_desc}), thus by mitigating bias for race, the prediction bias (MPE gap) for income has been reduced simultaneously.\\

\noindent Figures \ref{fig:conv_black} and \ref{fig:pac_black} plot MPE for black and non-black groups as well as the overall MAE corresponding to Table \ref{tab:result_bias_race}. As we increase the bias mitigation weight ($\gamma$), the prediction MPEs for the black population (denoted by the green bars) decrease considerably, indicating that with the bias mitigation loss function, the underestimation of TNC demand for the black population has been mitigated. \\
\FloatBarrier
\begin{table}[H]\centering 
  \caption{Fairness and accuracy comparisons with bias mitigation for race}
  \label{tab:result_bias_race} 
\resizebox{\textwidth}{!}{%
\begin{tabular}{lll|lll|lll}
\\[-1.8ex]\hline 
\hline \\[-1.8ex] 
                              & MAE   & MAPE  & MPE gap & Black  & Non-black & MPE gap  & Low-income & High-income \\
                              &       &       & (race)  & (MPE)  & (MPE)     & (income) & (MPE)      & (MPE)       \\
\hline \\[-1.8ex] 
\textit{Conv-LSTM Net:} &&&&&&&&      \\
$\gamma$ = 0  & 6.358 & 0.427 & 0.361   & 0.350  & -0.011    & 0.193    & 0.224      & 0.031       \\
$\gamma$ = 5  & 6.356 & 0.409 & 0.218   & 0.205  & -0.013    & 0.092    & 0.117      & 0.025       \\
$\gamma$ = 10  & 6.370 & 0.448 & 0.084   & 0.052  & -0.032    & 0.006    & 0.003      & -0.003      \\
\textit{SA-Net:} &&&&&&&&      \\
$\gamma$ = 0   & 6.279 & 0.407 & 0.272   & 0.235  & -0.037    & 0.127    & 0.131      & 0.004       \\
$\gamma$ = 5   & 6.259 & 0.405 & 0.235   & 0.212  & -0.024    & 0.103    & 0.118      & 0.015       \\
$\gamma$ = 10  & 6.228 & 0.401 & 0.111   & 0.124  & 0.012     & 0.066    & 0.088      & 0.022    \\ 
\hline 
\hline \\[-1.8ex] 
\multicolumn{8}{l}{\textit{Note:} $\gamma$ represents the bias mitigation weight}\\   
\end{tabular}%
} 
\end{table}
\FloatBarrier

\FloatBarrier
\FloatBarrier

\noindent Then, we apply the same bias mitigation strategy to mitigate the MPE gap between the low-income and high-income groups. In this case, $z$ denotes the proportion of low-income population in Equation \ref{equ:L_fairness}. The results for ``MPE gap (income)" in Table \ref{tab:result_bias_income} show that similar to the bias mitigation results for race, the MPE gaps decline as $\gamma$ increases when mitigating the income bias for both Conv-LSTM Net and SA-Net. Specifically, when $\gamma$ increases from 0 to 10, the MPE gap between the low-income and high-income groups decreases from 0.193 to -0.044 for Conv-LSTM Net, and decreases from 0.127 to 0.059 for SA-Net. The MPE gaps between two income groups are also plotted in Figure \ref{fig:conv_income} and Figure \ref{fig:pac_income}, where we can see that the de-biasing regularization especially works well to reduce the MPE gap in the Conv-LSTM Net. \\

\FloatBarrier
\begin{table}[H]\centering 
  \caption{Fairness and accuracy comparisons with bias mitigation for income}
  \label{tab:result_bias_income} 
\resizebox{\textwidth}{!}{%
\begin{tabular}{lll|lll|lll}
\\[-1.8ex]\hline 
\hline \\[-1.8ex] 
                              & MAE   & MAPE  & MPE gap & Black  & Non-black & MPE gap  & Low-income & High-income \\
                              &       &       & (race)  & (MPE)  & (MPE)     & (income) & (MPE)      & (MPE)       \\
\hline \\[-1.8ex] 
\textit{Conv-LSTM Net:} &&&&&&&&      \\
$\gamma$ = 0  & 6.358 & 0.427 & 0.361   & 0.350  & -0.011    & 0.193    & 0.224      & 0.031       \\
$\gamma$ = 5  & 6.338 & 0.414 & 0.281   & 0.276  & -0.005    & 0.117    & 0.161      & 0.044       \\
$\gamma$ = 10 & 6.295 & 0.428 & 0.002   & -0.030 & -0.032    & -0.044   & -0.054     & -0.010      \\
\textit{SA-Net:} &&&&&&&&      \\
$\gamma$ = 0   & 6.279 & 0.407 & 0.272   & 0.235  & -0.037    & 0.127    & 0.131      & 0.004       \\
$\gamma$ = 5   & 6.198 & 0.408 & 0.240   & 0.201  & -0.039    & 0.062    & 0.083      & 0.021       \\
$\gamma$ = 10  & 6.257 & 0.406 & 0.207   & 0.182  & -0.025    & 0.059    & 0.083      & 0.024  \\ 
\hline 
\hline \\[-1.8ex] 
\multicolumn{8}{l}{\textit{Note:} $\gamma$ represents the bias mitigation weight}\\   
\end{tabular}%
} 
\end{table}
\FloatBarrier

\noindent In addition, we find that the improving prediction fairness does not necessarily sacrifice prediction accuracy. The orange dots in Figure \ref{fig_chart} denote the MAEs produced by different models, which show that compared with no bias mitigation, only Conv-LSTM shows a slight increase in MAE when mitigating the racial bias (Figure \ref{fig:conv_black}). Under other circumstances, the application of bias mitigation actually also brings down MAE. Notably, when increasing the mitigation weight $\gamma$ for income from 0 to 5 for SA-Net, the prediction accuracy has been greatly improved (MAE=6.198) compared with the case when no bias mitigation is adopted (MAE=6.279). \\

\begin{figure}[!h]
\centering
\subfloat[Conv-LSTM Net: black]{\label{fig:mitigate_conv_min_black}\includegraphics[width=2in]{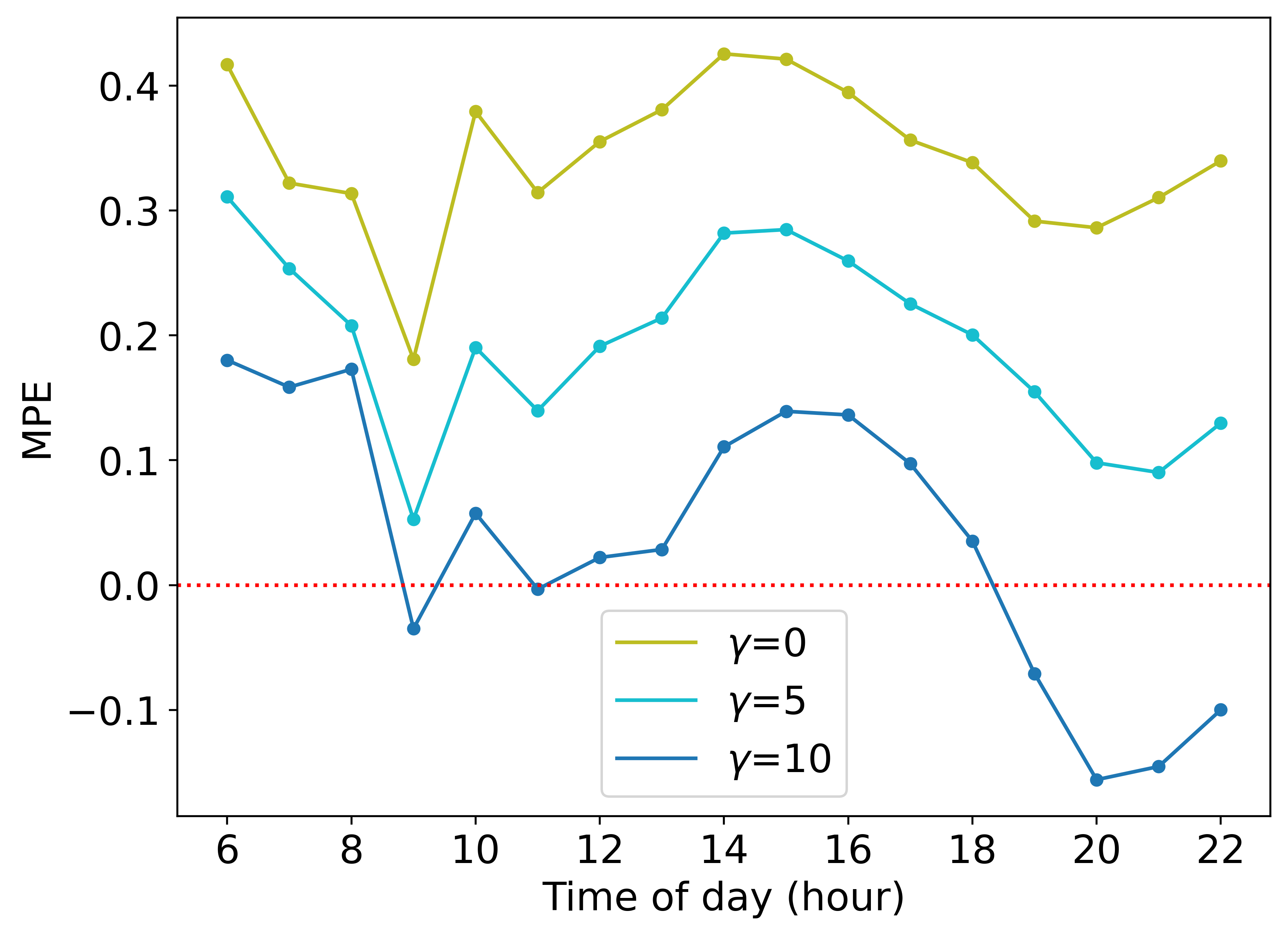}} 
\quad
\subfloat[Conv-LSTM Net: non-black]{\label{fig:mitigate_conv_maj_black}\includegraphics[width=2in]{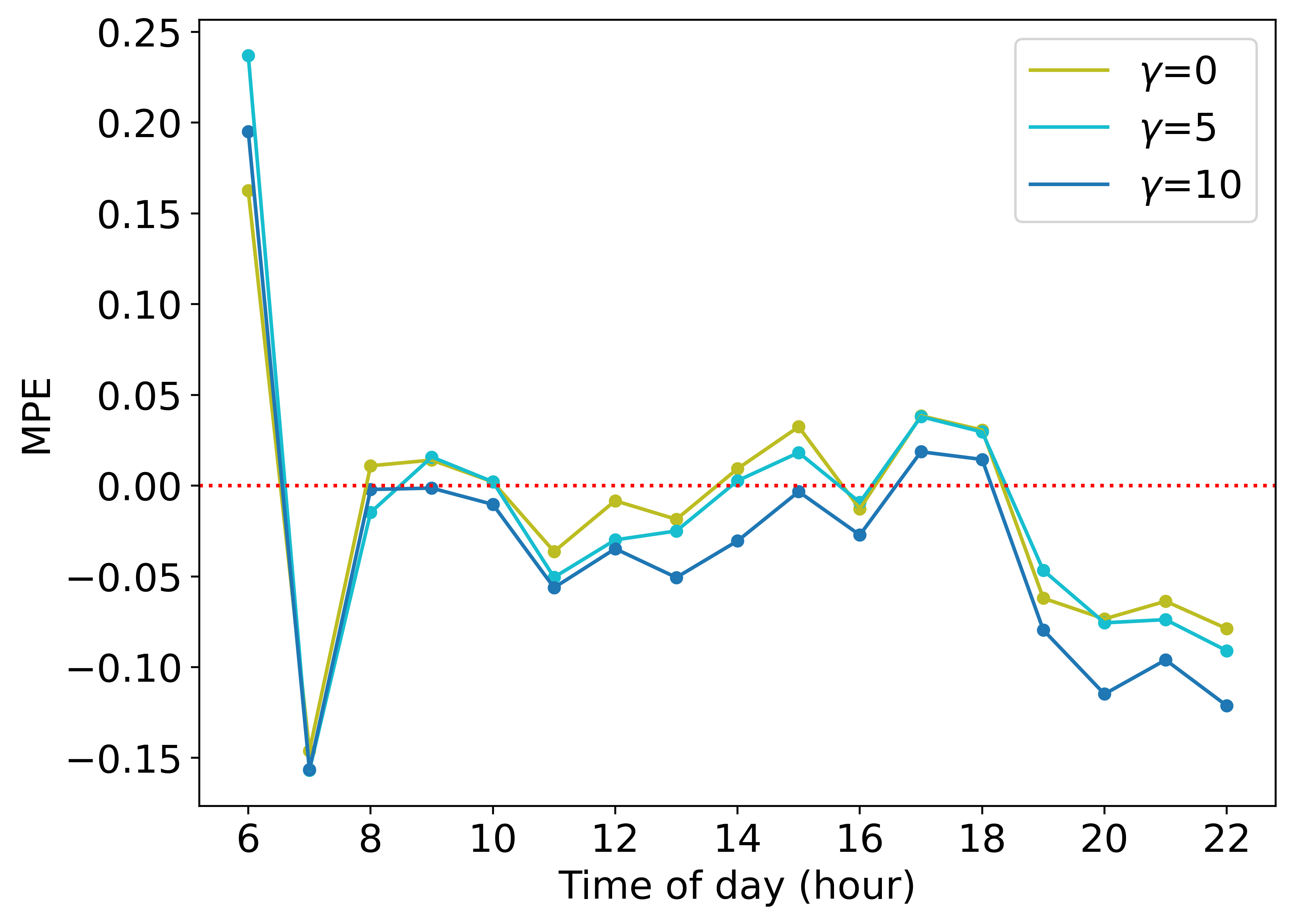}}
\quad
\subfloat[Conv-LSTM Net: MPE gap (black v.s. non-black)]{\label{fig:mitigate_conv_gap_black}\includegraphics[width=2in]{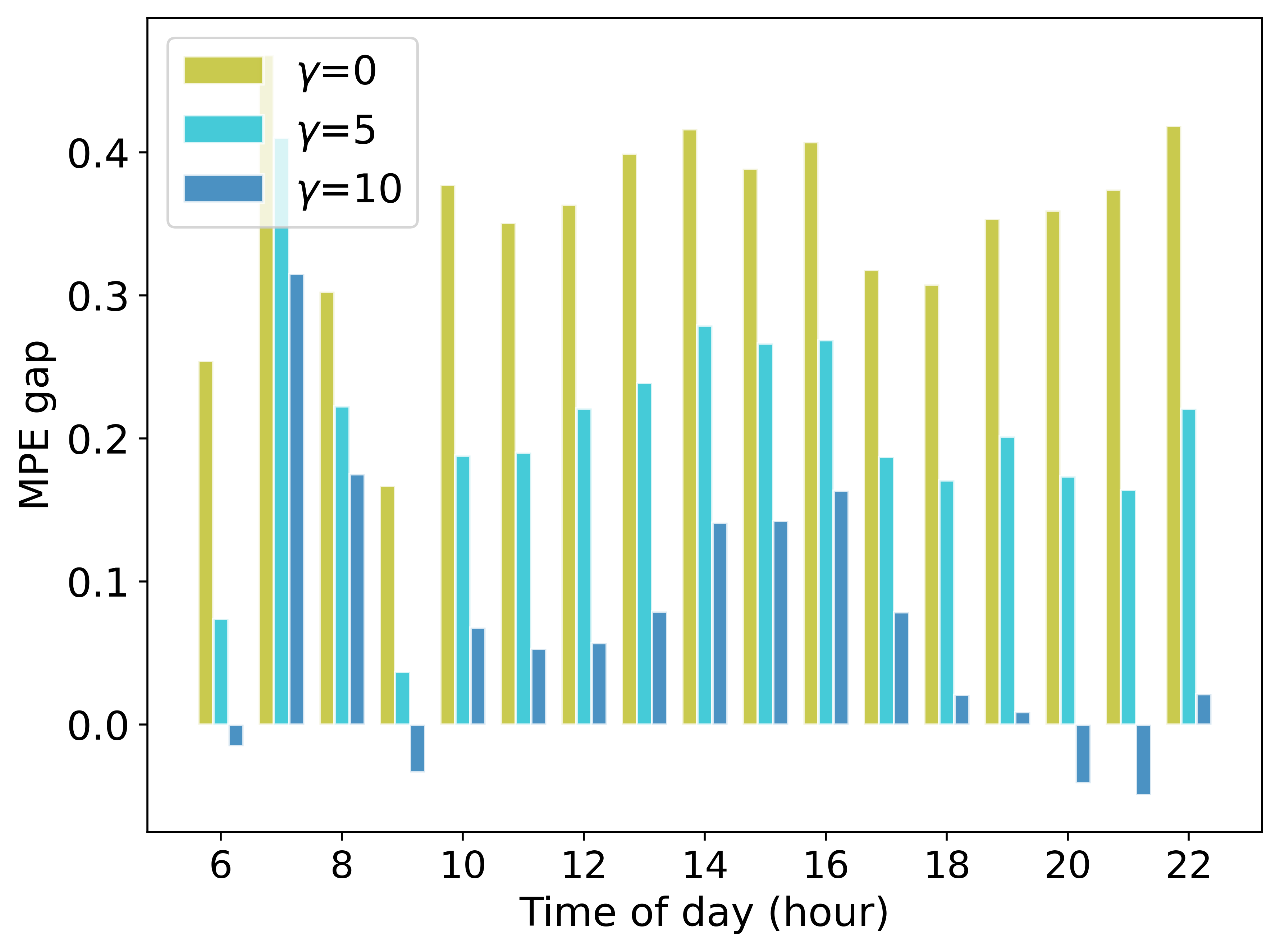}}\par\medskip

\subfloat[SA-Net: black]{\label{fig:mitigate_pac_OneMap_min_black}\includegraphics[width =2in]{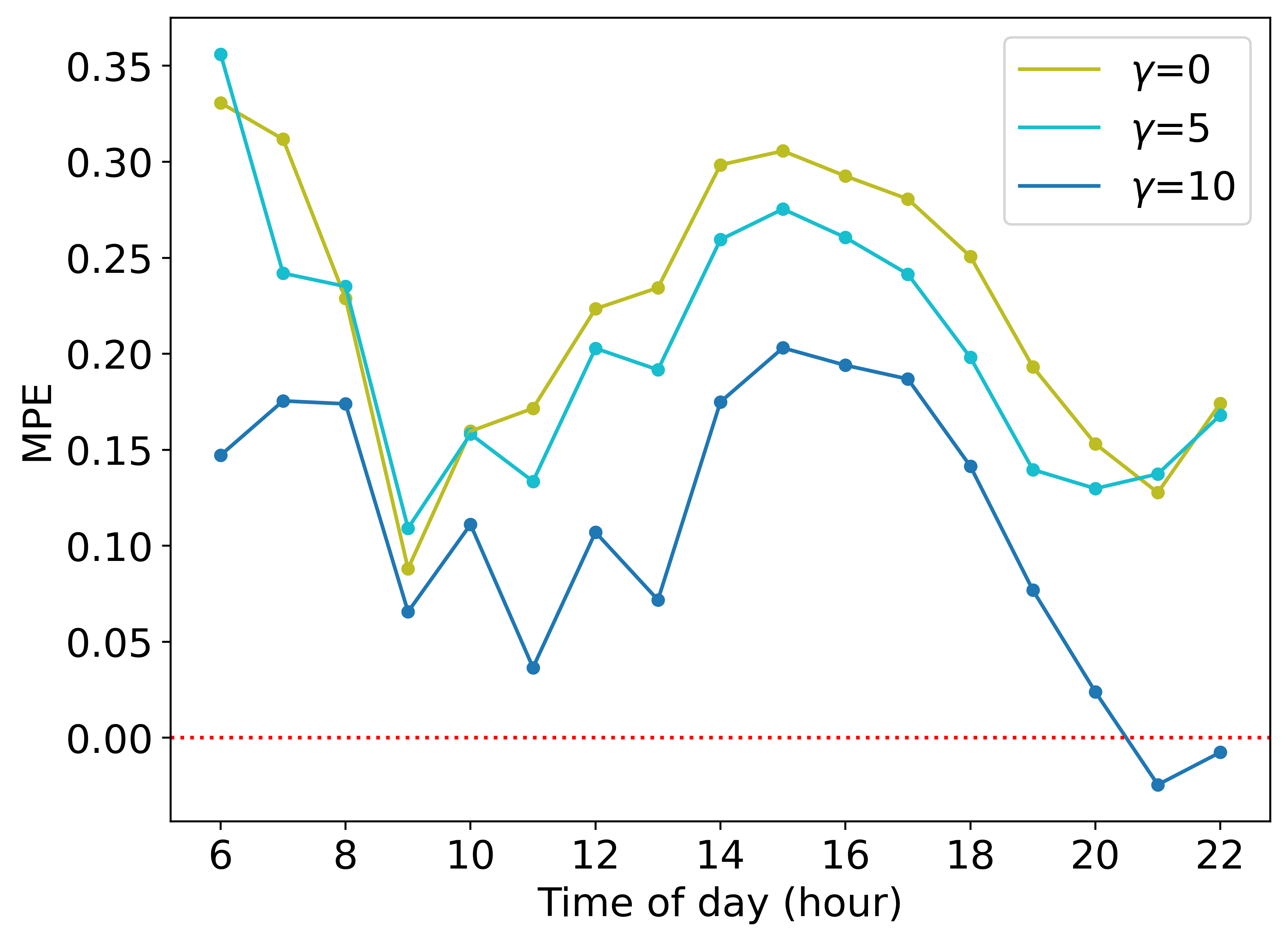}}
\quad
\subfloat[SA-Net: non-black]{\label{fig:mitigate_pac_OneMap_maj_black}\includegraphics[width =2in]{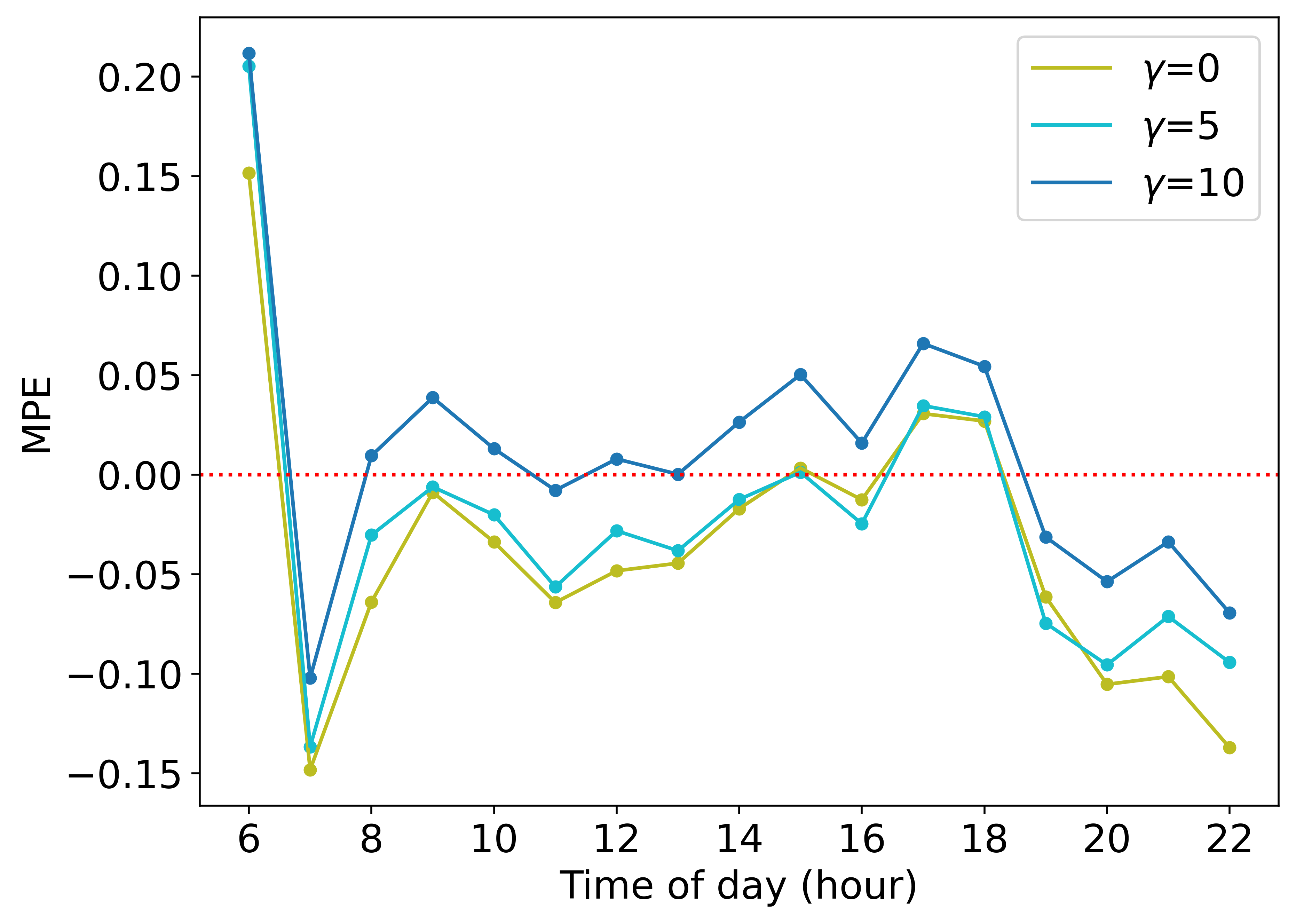}}
\quad
\subfloat[SA-Net: MPE gap (black v.s. non-black)]{\label{fig:mitigate_pac_OneMap_gap_black}\includegraphics[width =2in]{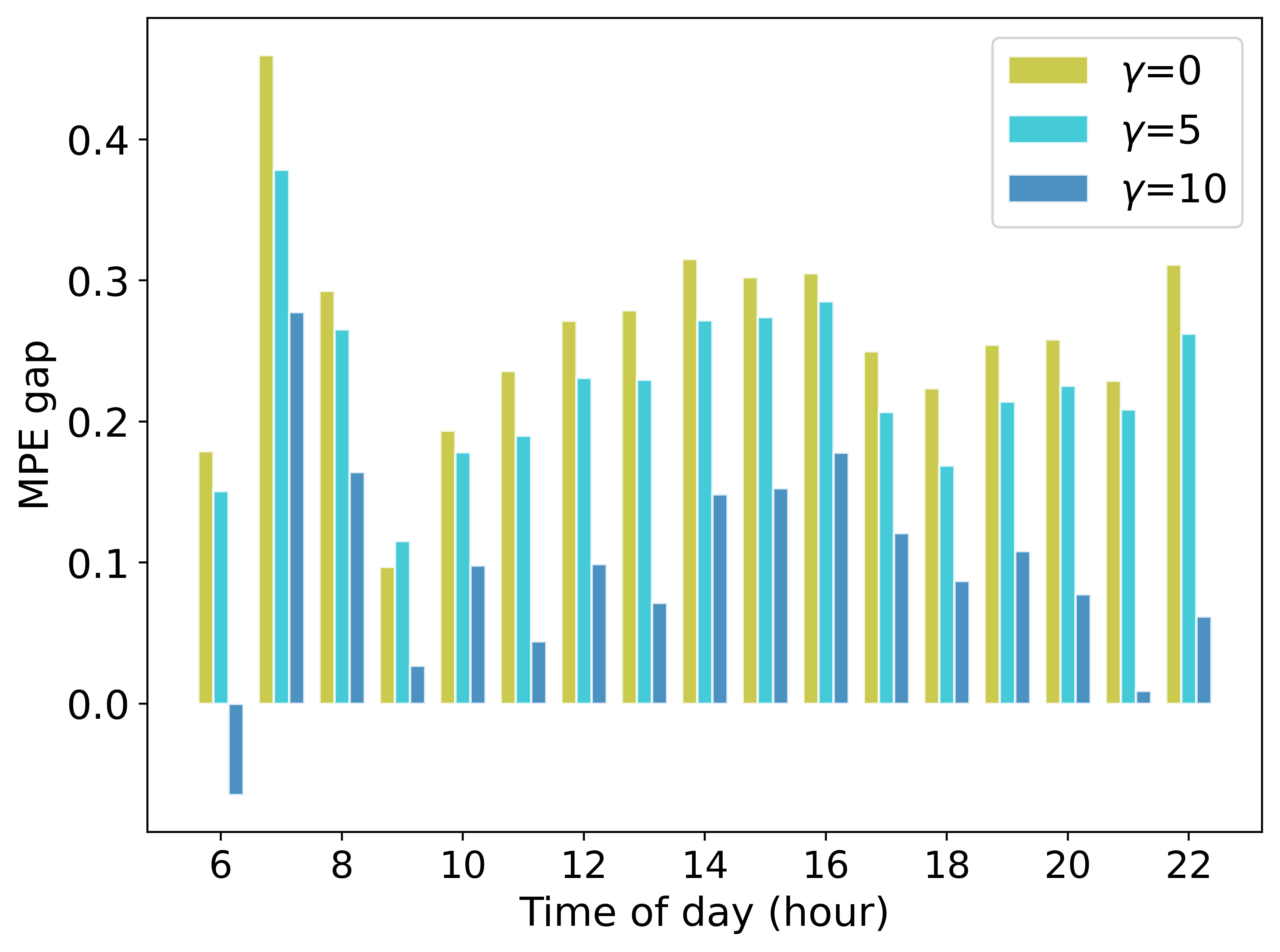}}
\caption{MPE for different racial groups with different mitigation weights ($\gamma$) by time of day} 
\label{fig_chart_mitigate} 
\end{figure}

\noindent We also examine the change of average MPE in different time of day with different bias mitigation strategies in Figure \ref{fig_chart_mitigate}. Figure \ref{fig:mitigate_conv_min_black} and \ref{fig:mitigate_pac_OneMap_min_black} show that by increasing the bias mitigation weight $\gamma$ from 0 to 5 and 10, the MPE  for the black communities consistently decreases in all times of day for both the Conv-LSTM Net and the SA-Net, and the bias mitigation effect is slightly stronger in the Conv-LSTM Net case. For the Conv-LSTM Net, we observe that when $\gamma=10$, the morning peak MPE decreases to around zero, and the MPE for the evening period (after 6 pm) becomes negative. On the contrary, Figure \ref{fig:mitigate_conv_maj_black} and \ref{fig:mitigate_pac_OneMap_maj_black} show that the effects of the bias mitigation method on the MPE for the non-black communities are relatively small. The increase of $\gamma$ is associated with a small drop of MPE in the Conv-LSTM Net case and a small rise of MPE in the SA-Net case. Figure \ref{fig:mitigate_conv_gap_black} and \ref{fig:mitigate_pac_OneMap_gap_black} plot the gaps in MPE between the black and non-black communities given by Conv-LSTM Net and SA-Net, which show that for all times of day, increasing the bias mitigation weight reduces the MPE gap between the black and non-black communities. All in all, our results suggest that our proposed bias mitigation strategy can significantly mitigate the travel demand underprediction issue for the black communities in all times of day with both the Conv-LSTM Net and the SA-Net, and can effectively reduce the prediction bias between the black and non-black groups.\\



\FloatBarrier
\FloatBarrier

\noindent In summary, our proposed de-biasing regularization method can considerably reduce the prediction bias measured by the MPE gap between the disadvantaged and the privileged groups for both Conv-LSTM Net and SA-Net. This gain in prediction fairness can be achieved while keeping the prediction accuracy high. For SA-Net, adopting bias mitigation can even increase prediction accuracy.\\


\subsubsection{5.4.3 Spatial patterns of errors}
\begin{figure}[!h]
\centering
\subfloat[No bias mitigation (Moran's I $=0.477$)]{\label{fig:no_mitig_map}\includegraphics[width=2in]{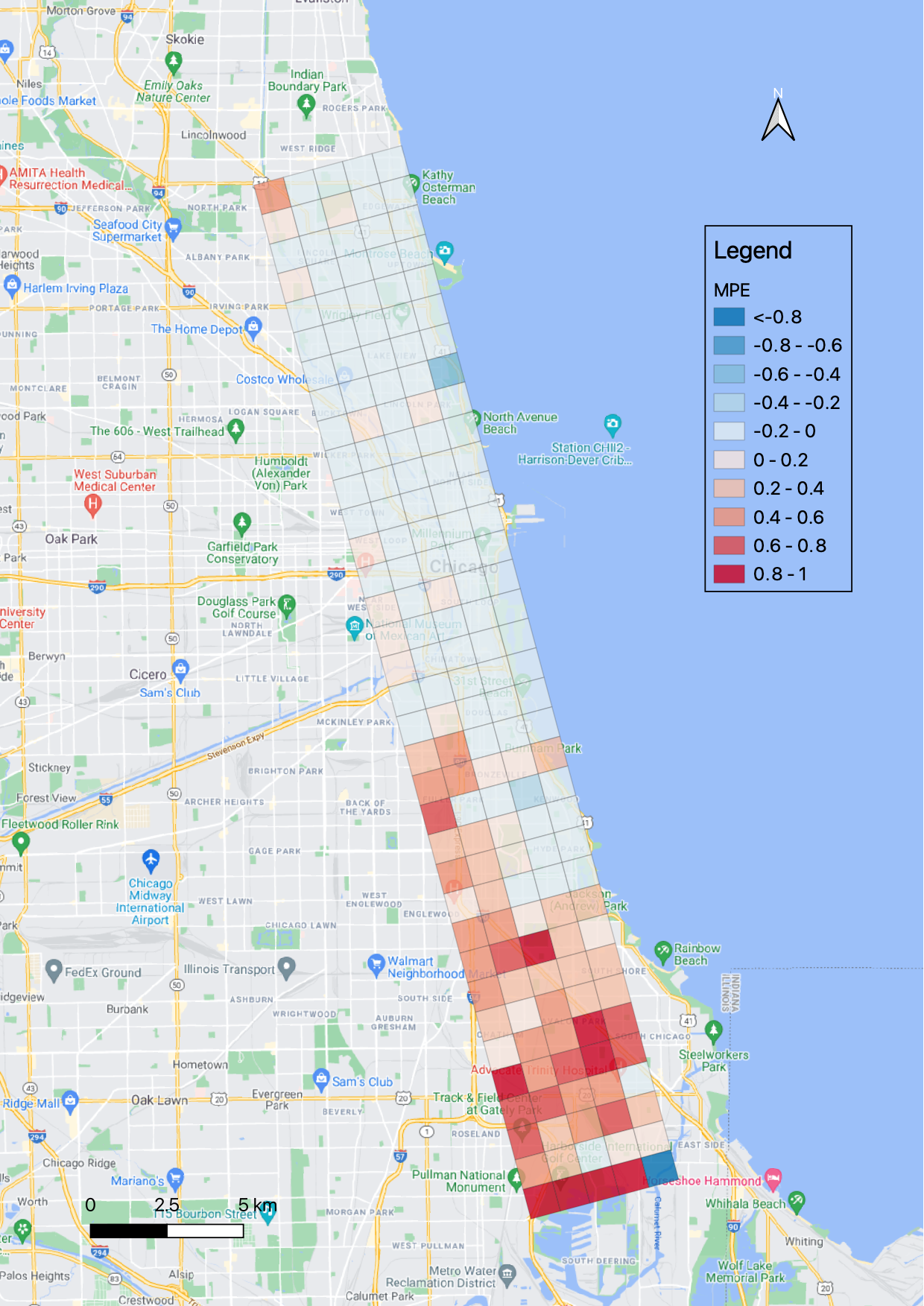}} 
\quad
\subfloat[Bias mitigation for race (Moran's I $=0.284$)]{\label{fig:black_mitig_map}\includegraphics[width=2in]{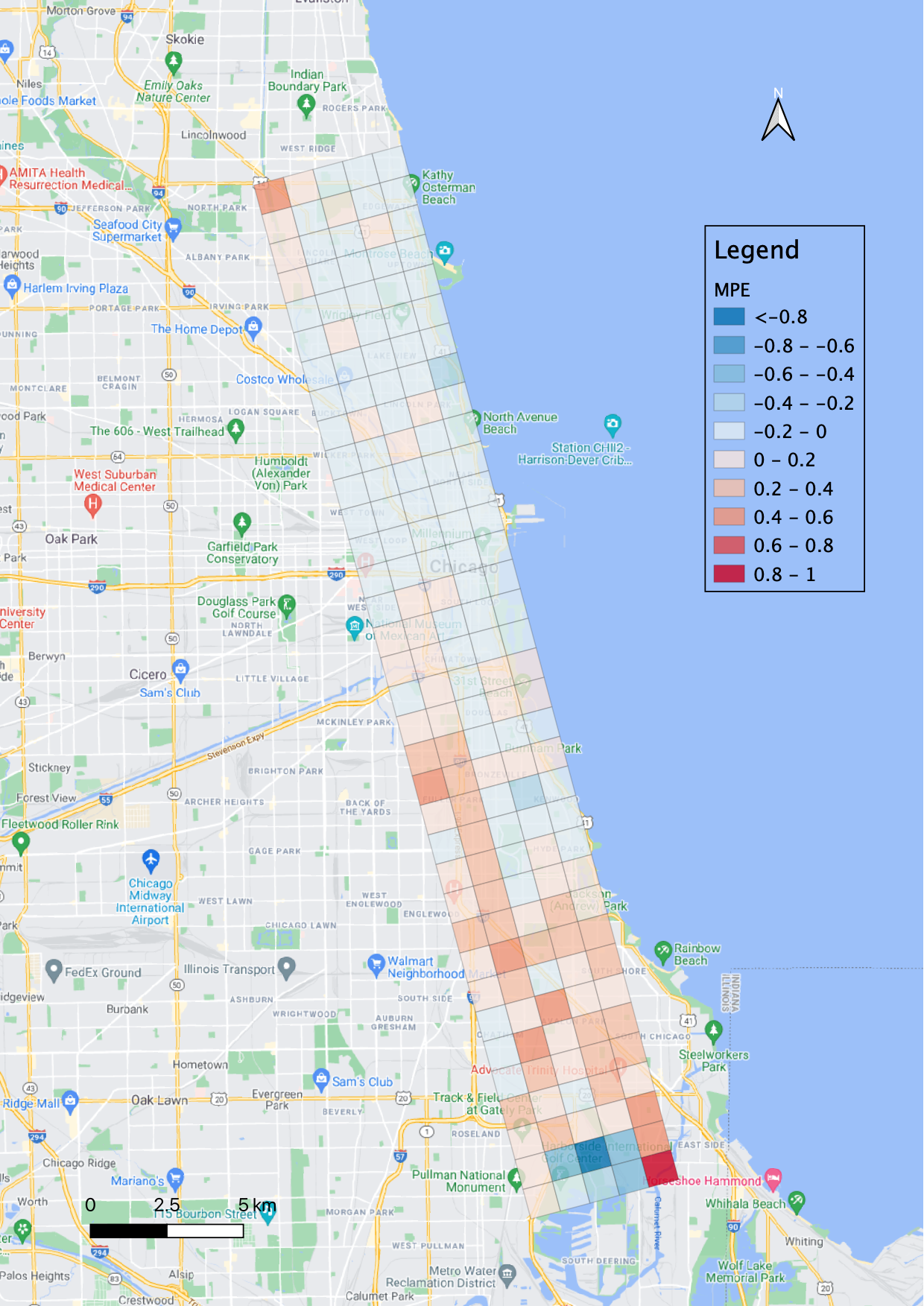}} 
\quad
\subfloat[Bias mitigation for income (Moran's I $=0.39$)]{\label{fig:income_mitig_map}\includegraphics[width =2in]{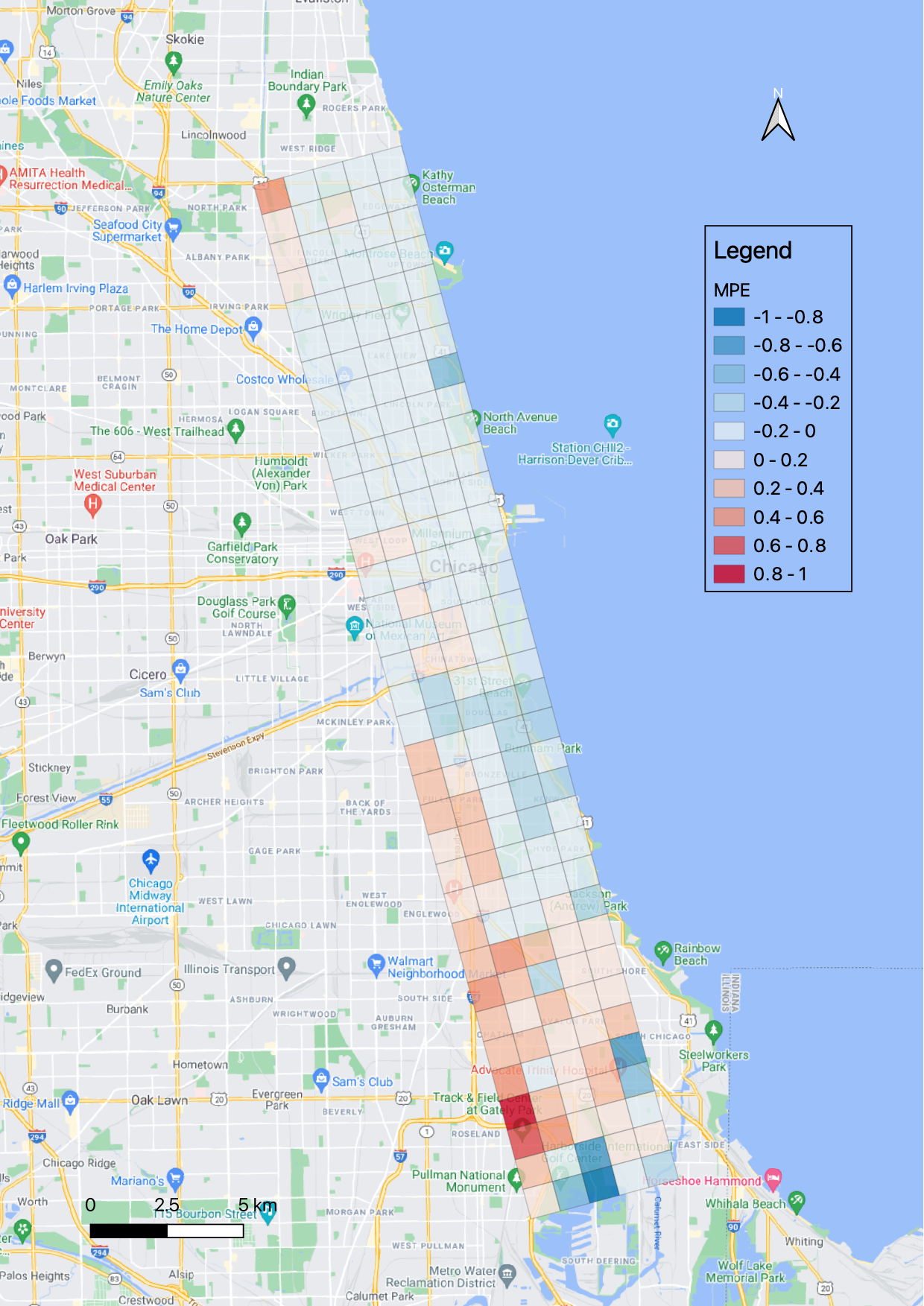}}
\caption{Spatial distributions of mean percentage errors using SA-Net for different bias mitigation strategies} 
\label{fig_map}
\end{figure}

\noindent To better understand the spatial heterogeneity of the prediction errors, we show in Figure \ref{fig_map} the spatial distributions of MPE using SA-Net for three prediction strategies: prediction with no bias-mitigation, with race bias mitigation and with income bias mitigation. Areas with positive MPE (indicating that the TNC demand has been underestimated) are denoted by the red color, whereas areas with negative MPE (indicating demand overestimation) are denoted by the blue color. Figure \ref{fig:no_mitig_map} shows that when no bias mitigation is adopted, the south side of the study area, which has greater populations of low-income and African-American people, suffers from severe demand underestimation. When we add the bias mitigation for race and income, the results in Figure \ref{fig:black_mitig_map} and Figure \ref{fig:income_mitig_map} show that the grid colors in the southern areas have become much lighter, and the colors of several areas in the south switch from red to blue, suggesting that the underestimation issue has been remarkably alleviated. \\

\noindent The Moran's I measure has been calculated for MPE in each scenario. Moran's Index, or simply Moran's I, is a spatial autocorrelation coefficient that has been widely used to measure the degree to which geographic events clustered in the study area. The results show that by applying bias mitigation for race and income, the Moran's I value reduces from 0.477 to 0.284 and 0.39, respectively, which demonstrates that our bias mitigation method can help reduce spatial clustering of MPE.

\FloatBarrier
\FloatBarrier

\section{6. Discussion and conclusions}
\label{sec:conclusion}
Fairness has long been a critical concern in transportation studies. However, the prediction fairness issue in spatial-temporal travel demand forecasting has been neglected in previous literature. In this paper, we propose a two-pronged approach to enhance fairness in TNC demand forecasting, and test the effectiveness of our innovative method on Chicago's TNC data.\\

\noindent First, though previous studies have shown that there is significant difference in travel characteristics across demographic groups \cite{giuliano2003travel,gao2021segregation,zheng2022gender}, most spatial-temporal research failed to account for the socio-demographic heterogeneity in travel demand predictions. We propose a new model structure SA-Net that can flexibly capture the variations of spatial correlations across different socio-demographic groups. The experimental result shows that our proposed SA-Net can both improve the overall prediction accuracy and promote fairness by delivering more accuracy gain for the disadvantaged (i.e. black and low-income) communities while not harming the performance of the privileged (i.e. non-black and high-income) communities.\\

\noindent Second, we find that previous solutions to spatial-temporal travel demand prediction problems tended to underestimate demand for the low-demand regions. This is because high errors in low-demand areas will significantly impact MAPE, thus optimizing MAPE would likely underestimate demand in these areas. To tackle this issue, we use the mean percentage error gap to measure prediction fairness, and propose a novel regularization method to mitigate the bias between the disadvantaged and privileged groups through disentangling the correlation between the sensitive attribute and the mean percentage error. Our experimental results show that the new algorithm can effectively mitigate prediction bias for both the traditional Conv-LSTM Net and the new SA-Net. It can also protect the disadvantaged regions against systematic underestimation. Our results also show that the new method can improve prediction fairness while retaining high prediction accuracy.\\

\noindent Overall, we argue that the prediction bias issue revealed in this work should attract the attention of the researchers and policy makers, because if the travel demand in the disadvantaged neighborhoods is systematically underpredicted, we may fail to provide enough TNC services to these communities, and the limited services will in turn lead to further decrease of the ridership, which will eventually lead to a negative feedback loop. The method proposed in this study has been proven to be capable of tackling this prediction bias issue and promoting both accuracy and fairness.\\

\noindent We identify several future research directions worth investigating. First, this paper evaluates fairness in travel demand prediction and demonstrates the utility of the de-biasing mitigation method on Conv-LSTM Net and the SA-Net. However, the proposed fairness evaluation metrics and the bias mitigation method are widely applicable. They can also be applied to other spatial-temporal deep learning networks such as the spatial-temporal residual networks ST-ResNet \cite{zhang2017deep} and RSTN \cite{guo2020residual}. Second, this study aims to implement fair predictions for on-demand ride service. However, our proposed fairness-enhancing method should also work well for other spatial-temporal settings, such as bikeshare demand prediction, public transport demand prediction and crime incidents prediction. Future research can test the performance of the proposed method on various downstream applications. Third, we test our method on Chicago's TNC data as the real-world application. Future work can study the transferability of our method to other applications or cities.


\newpage
\bibliographystyle{trb}
\bibliography{main}
\nolinenumbers
\end{document}